\documentclass[runningheads]{llncs}

\usepackage{eccv}

\usepackage{eccvabbrv}

\usepackage{graphicx}
\usepackage{booktabs}
\usepackage[inline]{enumitem}
\usepackage{multirow}
\usepackage{makecell}
\usepackage{xcolor,colortbl}

\usepackage[accsupp]{axessibility}  % Improves PDF readability for those with disabilities.

\usepackage{xargs}                      % Use more than one optional parameter in a new commands
\usepackage[colorinlistoftodos,prependcaption,textsize=tiny]{todonotes}
\newcommandx{\unsure}[2][1=]{\todo[linecolor=red,backgroundcolor=red!25,bordercolor=red,#1]{#2}}
\newcommandx{\change}[2][1=]{\todo[linecolor=blue,backgroundcolor=blue!25,bordercolor=blue,#1]{#2}}
\newcommandx{\info}[2][1=]{\todo[linecolor=OliveGreen,backgroundcolor=OliveGreen!25,bordercolor=OliveGreen,#1]{#2}}
\newcommandx{\improvement}[2][1=]{\todo[linecolor=Plum,backgroundcolor=Plum!25,bordercolor=Plum,#1]{#2}}
\newcommandx{\thiswillnotshow}[2][1=]{\todo[disable,#1]{#2}}

\usepackage{hyperref}

\usepackage{orcidlink}

\begin{document}

\title{GazeXplain: Learning to Predict Natural Language Explanations of Visual Scanpaths} 

\titlerunning{Learning to Predict Natural Language Explanations of Visual Scanpaths}

\author{Xianyu Chen\orcidlink{0000-0002-9027-3920} \and
Ming Jiang\orcidlink{0000-0001-6439-5476} \and
Qi Zhao\orcidlink{0000-0003-3054-8934}}

\authorrunning{X.~Chen et al.}

\institute{University of Minnesota, Minneapolis MN 55455, USA\\
\email{\{chen6582,mjiang\}@umn.edu, qzhao@cs.umn.edu}}

\maketitle

\begin{abstract}
   While exploring visual scenes, humans' scanpaths are driven by their underlying attention processes. Understanding visual scanpaths is essential for various applications. Traditional scanpath models predict the where and when of gaze shifts without providing explanations, creating a gap in understanding the rationale behind fixations. To bridge this gap, we introduce GazeXplain, a novel study of visual scanpath prediction and explanation. This involves annotating natural-language explanations for fixations across eye-tracking datasets and proposing a general model with an attention-language decoder that jointly predicts scanpaths and generates explanations. It integrates a unique semantic alignment mechanism to enhance the consistency between fixations and explanations, alongside a cross-dataset co-training approach for generalization. These novelties present a comprehensive and adaptable solution for explainable human visual scanpath prediction. Extensive experiments on diverse eye-tracking datasets demonstrate the effectiveness of GazeXplain in both scanpath prediction and explanation, offering valuable insights into human visual attention and cognitive processes.
\end{abstract}

%-------------------------------------------------------------------------
\begin{figure*}[tb]
\centering
\includegraphics[width=0.9\linewidth]{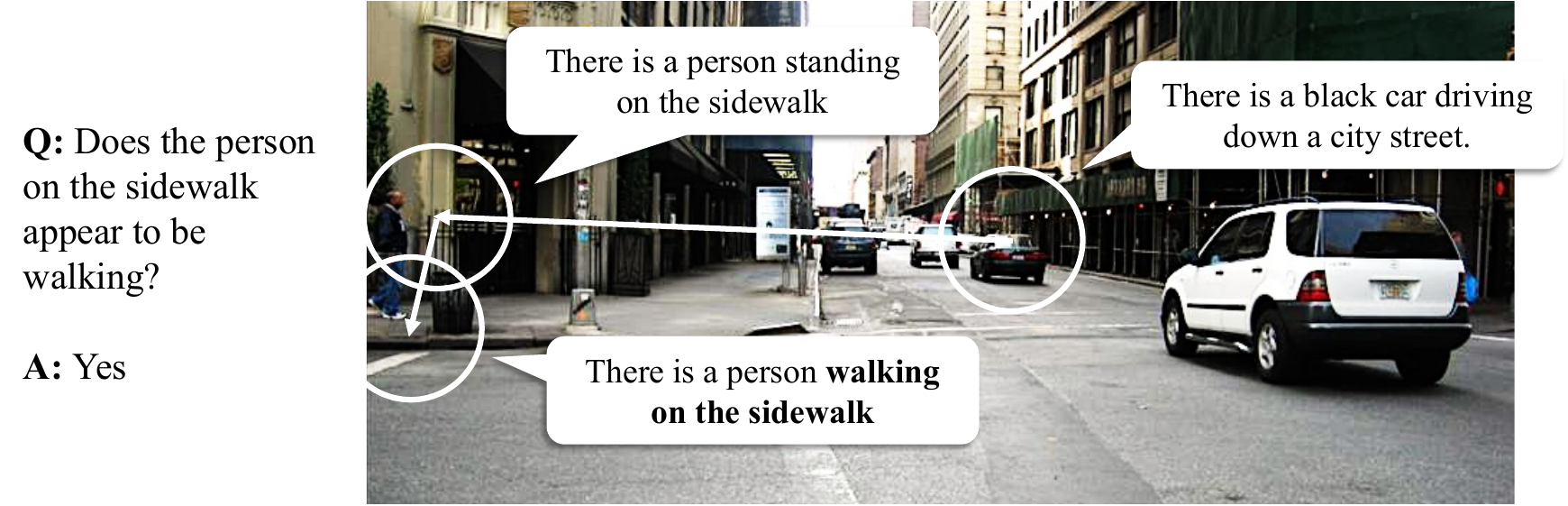}
\caption{This example reveals how observers strategically investigate a scene to find out if the person is walking on the sidewalk. Fixations (circles) start centrally, locating a driving car, then shifting to the sidewalk to find the person, and finally looking down to confirm if the person is walking. By annotating observers' scanpaths with detailed explanations, we enable a deeper understanding of the ``what'' and ``why'' behind fixations, providing insights into human decision-making and task performance.}
\label{fig:teaser}
\end{figure*}
%-------------------------------------------------------------------------

\section{Introduction}
\label{sec:intro}

Picture yourself driving through a bustling city at dusk, with your eyes scanning the surroundings for critical details like pedestrian crossings, brake lights, and turn signals. These seemingly random glances are guided by an internal dialogue questioning your environment. As depicted in \cref{fig:teaser}, when determining if a person on the sidewalk is standing or walking, our gaze naturally shifts from the car ahead to the sidewalk. We may fixate on their upper body to start with, and then move downward to assess their movement. Understanding this implicit language of gaze and translating it into explicit explanations, such as whether we correctly deduced the person's movement or overlooked subtle details, holds significant potential for enhancing human-machine interaction.

Research on human attention modeling builds upon decades of study in psychology and cognitive science, aiming to understand how humans allocate their attention to visual stimuli~\cite{xun:2015:salicon,ming:2015:salicon}. Recent studies have shifted from static fixation distribution modeling to dynamic gaze patterns, known as scanpaths~\cite{olivier:2015:saccadicmodel,laurent:1998:visualattention,xianyu:2021:vqa,sounak:2023:gazeformer}. 
Current scanpath models excel at tracking visual exploration trajectories, predicting ``when'' and ``where'' people shift their attention. However, scanpath prediction models fall short of explicitly explaining the ``what'' and ``why'' -- the insights behind each fixation. This lack of explainability hampers the understanding and potential applications of these models in real-world scenarios.

To bridge this explainability gap, we introduce GazeXplain, a novel study that goes beyond predicting where people look; it demands models to explain them in natural language, weaving a narrative thread that connects fixations to their underlying meaning. Particularly, GazeXplain features several key distinctions from existing scanpath prediction methods: (1) We annotate ground-truth explanations for scanpaths over diverse eye-tracking datasets. These annotations build a strong foundation for modeling scanpath explanation, unlocking explainable methods that understand user attention in applications. (2) We introduce a general model architecture with an attention-language decoder jointly predicting scanpaths and natural language explanations. (3) We present a novel semantic alignment mechanism that promotes consistency between the vision and language modalities, guiding the model toward generating explanations that faithfully reflect the fixated visual information. (4) While existing models target single task-specific datasets, such as free-viewing, object search, or visual question answering (VQA), we generalize scanpath prediction and explanation with a cross-dataset co-training technique, overcoming data and task-specific biases.

In summary, the contributions of this paper are outlined as follows:
\begin{enumerate}
  \item We introduce a novel task aiming to jointly predict and explain scanpaths, offering a deeper semantic understanding of what people look.
  \item We annotate ground-truth explanations on three public eye-tracking datasets, providing detailed fixation-level explanations.
  \item We propose a general model architecture with an attention-language decoder that jointly predicts scanpath and explanation. It incorporates a novel semantic alignment mechanism for consistent fixation-explanation alignment, along with cross-dataset co-training for enhanced generalizability.
  \item Comprehensive experiments across various datasets demonstrate GazeXplain's effectiveness in generating accurate scanpaths and explanations, highlighting the importance of explanation prediction, semantic alignment, and cross-dataset co-training on model performance.
\end{enumerate}

\section{Related Work}
\label{sec:related_works}

\textbf{Visual Scanpath Prediction.} 
Understanding human visual attention requires insight into the dynamic sequence of eye fixations. While static saliency prediction has been extensively studied~\cite{matthias:2016:deepgaze,marcella:2018:sam,xun:2015:salicon,camilo:2020:umsi,souradeep:2022:agdf,sen:2020:eml,shi:2023:personalsaliency,bahar:2023:tempsal}, dynamic scanpath prediction remains relatively underexplored due to its complexity influenced by various factors. Early studies employed heuristics or statistical priors to generate scanpaths~\cite{dirk:2000:gazeshift,laurent:2000:visualattention,laurent:2001:visualattention,laurent:1998:visualattention,dirk:2006:attention}, while recent models leverage machine learning techniques, including supervised learning~\cite{matthias:2016:deepgaze,mengyu:2023:scanpath,sounak:2023:gazeformer,zhibo:2023:hat,wanjie:2019:iorroi,zhenzhong:2018:iorroi-lstm,matthias:2022:deepgaze} and reinforcement learning~\cite{xianyu:2021:vqa,zhibo:2020:cocosearch,zhibo:2022:targetabsent}, achieving promising results~\cite{xianyu:2021:vqa,zhibo:2020:cocosearch,zhibo:2022:targetabsent,mengyu:2023:scanpath,sounak:2023:gazeformer,xianyu:2024:individualscanpath,peizhao:2023:uniar}. However, these methods lack interpretability and struggle to explain the predicted fixations. Our method, GazeXplain, stands out in two aspects: Firstly, it generates natural language explanations for predicted fixations, going beyond mere scanpath prediction. Secondly, it ensures generalizability across visual tasks by training on a combination of datasets. This improved explainability and generalizability represent significant advancements in understanding human visual attention processes.

\textbf{Explanations.}
Automatic reasoning and explanation~\cite{lisa:2021:explanation} initially rely on rules or predefined templates to explain medical diagnosis~\cite{edward:1975:reasonmedicine}, simulator actions~\cite{chad:2005:training,mark:2006:xai,michael:1999:xai,lewis:1994:xai} and robot movements~\cite{meghann:2012:xra} \etc. Recent explanation models explored deep learning-based natural language generation, with successful applications in producing natural language justifications for object classification~\cite{lisa:2016:explanation,kosuke:2022:fewshoweai,lisa:2018:counterfactual}, visual reasoning~\cite{shi:2022:rex,dong:2018:explanation,qing:2018:vqae,jialin:2019:explanationvqa,ana:2020:visualreasoning,radhika:2021:beyondvqa,jialin:2019:scvqa}, recommendation systems~\cite{hanxiong:2019:recommendation}, and sentiment analysis~\cite{zunwang:2021:knowledgeguide}, \etc.
Different from these studies, we for the first time explore natural language explanations of eye-tracking data to facilitate a deeper understanding of human visual behaviors. Our proposed GazeXplain model simultaneously predicts scanpaths and explanations, establishing a direct semantic connection to jointly improve the scanpath prediction and explanation accuracy.

\textbf{Vision and Language Models.} 
GazeXplain is inspired by the success of deep vision-language models~\cite{iro:2019:unsupervised-imagecaption,kelvin:2015:image-caption,dongjin:2019:scarcesupervised-imagecaption,hamed:2017:attentionimagecaptioning,yang:2019:unsupervised-imagecaption,xianyu:2021:anoc,jinhui:2022:visualhow,xianyu:2024:sgan,xianyu:2021:sd-fsic}. These models, trained on multimodal image and language datasets~\cite{xinlei:2015:mscoco,peter:2014:flickr30k,jinhui:2022:visualhow}, are able to generate fluent and accurate descriptions of visual information. The recent advent of transformer architecture~\cite{ashish:2017:transformer,lun:2019:AoA,marcella:2019:meshed-memory} marked a significant breakthrough, providing a robust framework for handling intricate relationships and long-range dependencies. This advancement facilitated the development of large-scale vision-language models that excel in translating visual information into natural language descriptions~\cite{junnan:2022:blip,junnan:2023:bilp2,haotian:2023:vit,haotian:2023:llava}. While these models have achieved impressive results in characterizing vision-language features, scanpath models haven't fully leveraged this capability to enhance human attention prediction. Unlike existing scanpath models, GazeXplain builds upon the strengths of vision-language models, incorporating explainability in scanpath prediction. By leveraging the capabilities of vision-language models for both ground-truth annotations and language modeling, GazeXplain deciphers the attention and reasoning behind fixations, bridging the gap between visual attention and language understanding.

\section{GazeXplain}
\label{sec:methodology}
Human visual attention is a complex interplay across multiple visual features and cognitive factors~\cite{laurent:2001:visualattention,jeremy:2017:fivefactors,antonio:2006:guidance} (\eg, low-level contrasts, objects, semantics, goals, and prior knowledge, \etc) However, existing deep learning-based scanpath models lack transparency in explaining how different factors influence their predictions. Our work tackles this challenge through novel dataset construction and modeling approaches: (1) We annotate new scanpath explanations based on existing eye-tracking datasets, offering ground-truth explanations for fixations across diverse tasks like VQA, free-viewing, and search. (2) We propose the first scanpath prediction and explanation model generating natural language explanations alongside predicted scanpaths, featuring novel techniques including attention-language decoder, semantic alignment, and cross-dataset co-training.

\subsection{Data}
\label{sec:dataset}
 We propose data annotation to offer ground-truth explanations for fixations across various eye-tracking datasets. Compared to previous image-to-language datasets, it has two key distinctions: (1) We present the first natural-language annotations on scanpaths, offering explanations for each specific fixation within the scanpath, rather than image-wise descriptions such as image captioning~\cite{xinlei:2015:mscoco,peter:2014:flickr30k} and visual storytelling~\cite{tinghao:2016:vist}. This granular level of detail offers deeper insights into the cognitive process behind each fixation. (2) While most image-to-language datasets focus on specific tasks, ours comprise a wider range of visual tasks, including free-viewing~\cite{juan:2014:osie}, object search~\cite{zhibo:2020:cocosearch,zhibo:2022:targetabsent}, and VQA~\cite{shi:2020:air}. This ensures the diversity of explanations, allowing models to be co-trained across multiple datasets to enhance their generalizability.

%-------------------------------------------------------------------------
\begin{figure*}[tb]
\centering
\includegraphics[width=\linewidth]{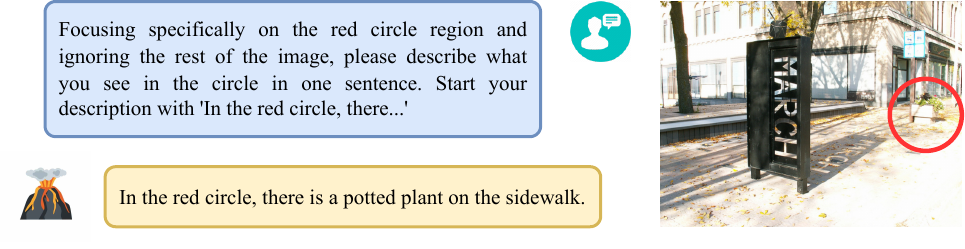}
\caption{LLaVA generates the ground-truth explanation for each fixation using an input image with a red circle marking the fixation. The model's response provides information within the marked area, serving as a basis for further refinement.}
\label{fig:vlm_demonstration}
\end{figure*}
%-------------------------------------------------------------------------

While the manual annotation of fixation-level explanations is subjective and time-consuming, we employ a novel semi-automated approach, leveraging the power of large vision-language models to efficiently generate accurate explanations for every eye fixation. \cref{fig:vlm_demonstration} illustrates our paradigm for annotating explanations. We utilize LLaVA-1.6~\cite{haotian:2023:llava} (with a Mistral-7B~\cite{albert:2023:mistral} base language model) for its renowned ability to understand and describe visual information. We combine visual and language prompts to guide the model's description: 
Firstly, we generate a visual prompt by enclosing each fixation within a red circle~\cite{aleksandar:2023:visualprompt}, mirroring the size of the human fovea~\cite{jeffrey:2022:gaze} (\ie, a diameter of 5 degrees), thereby directing LLaVA's attention to the fixated region. Complementing this, we crafted a language prompt that instructs LLaVA to describe the image information within the circled area in one sentence (see \cref{fig:vlm_demonstration}). These prompts guide LLaVA to generate concise and contextually relevant descriptions centered solely on the fixation. Preliminary evaluations have demonstrated the effectiveness of this prompting technique compared to alternative methods, such as describing multiple fixations simultaneously or generating foveated images as prompts, where the proposed one avoids issues related to information overload of multiple explanations or the complexity of computing foveated images.
Finally, these generated descriptions are combined in the order of fixations to describe the full scanpath, enabling the extraction of meaningful insights into the dynamic shift of attention.

While LLaVA's capabilities are impressive, it may exhibit limited robustness in handling noisy or ambiguous visual inputs, such as small objects, text, or complex scenes with cluttered backgrounds. Therefore, manual quality control remains crucial for ensuring accuracy and objectivity. To improve the data quality, we review and revise generated explanations based on the following criteria: Firstly, any reference to the red circle is eliminated to ensure that descriptions accurately reflect the information of the fixated regions. Secondly, for consistency and readability across datasets, the generated descriptions are maintained within a specific length (\eg, 5-20 words), facilitating subsequent analysis and interpretation. Thirdly, in images containing English text, the text recognition is manually verified and corrected. Finally, to ensure the consistency of explanations of fixations on the same object or region, we apply MeanShift~\cite{dorin:2002:meanshift} clustering to fixation positions and manually correct semantically different explanations in each cluster without sacrificing linguistic diversity. This quality control process enhances the overall accuracy, objectivity, and reliability of the annotations, mitigating potential errors introduced by automated processes.

%-------------------------------------------------------------------------
\begin{table*}[tb]
\caption{Statistics of the eye-tracking datasets with annotated explanations.}
\label{table:dataset-statistic}
\centering
\resizebox{1\textwidth}{!}{
{
\begin{tabular}{lcccccc}
 \toprule
  Dataset & Task & Images  & Scanpaths & Length of Scanpath &  Words per Fixation & Words per Scanpath\\
  \midrule
  AiR-D & VQA  & 987  & 13,903 & $10.17\pm2.23$ & $10.79\pm3.46$ & $109.81\pm31.27$\\
  OSIE & Free Viewing & 700 &  10,500 & $9.36\pm1.88$ & $11.43\pm3.99$ & $107.07\pm31.26$\\
  COCO-Search18 TP & Object Search  & 3,101 &   30,998 & $3.48\pm1.82$ & $9.84\pm3.14$ & $34.28\pm20.55$\\
  COCO-Search18 TA & Object Search & 3,101 & 31,006 & $5.86\pm4.07$ & $10.61\pm3.45$ & $62.21\pm45.85$\\
\bottomrule
\end{tabular}
}
}
\end{table*}
%-------------------------------------------------------------------------

By leveraging the combined strengths of LLaVA and human expertise, we annotat ground-truth explanations for three different eye-tracking datasets: AiR-D~\cite{shi:2020:air}, OSIE~\cite{juan:2014:osie}, as well as COCO-Search18~\cite{zhibo:2020:cocosearch} including target-present (TP) and target-absent (TA) subsets. As shown in \cref{table:dataset-statistic}, this results in a rich collection of natural-language explanations annotated on 7,004 images and 86,407 fixations across diverse visual tasks. The explanations are concise, with lengths falling within $10.66\pm3.54$ words each. 
The AiR-D dataset, involving question-answering scenarios, exhibited a range of explanation lengths (\ie, 10.79 per fixation), likely reflecting the varied complexity of the questions and corresponding fixations. Explanations for free-viewing tasks in OSIE tended to be slightly longer (\ie, 11.43 words per fixation) compared to search-oriented tasks in COCO-Search18 (\ie, 10.33 words per fixation). This aligns with the inherent differences in information processing during free exploration versus focused object search. Overall, the annotated explanations offer a valuable resource for researchers studying visual attention and its connection to language.

\subsection{Model}
\label{sec:architecture}

The core challenge in scanpath explanation is the mapping ambiguity: translating brief fixations with limited context into clear natural language descriptions. This difficulty stems from inherent subjectivity in visual perception and the lack of explicit semantic meaning behind each fixation. To address this, GazeXplain presents a three-fold solution (see \cref{fig:architecture}): 

%-------------------------------------------------------------------------
\begin{figure*}[tb]
\centering
\includegraphics[width=1.0\linewidth]{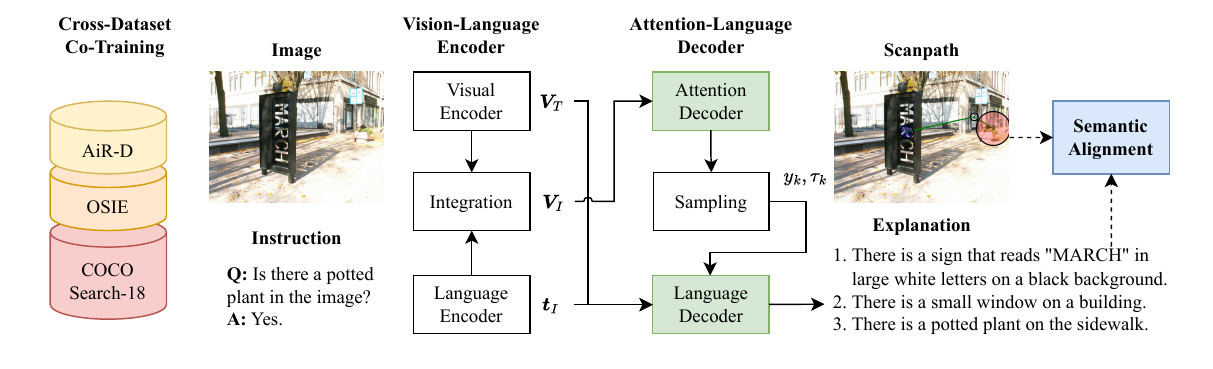}
\caption{GazeXplain's architecture consists of a general vision-language encoder and a novel attention-language decoder. The decoder outputs an explanation for each fixation in the predicted scanpath, with a semantic alignment mechanism facilitating the semantic consistency between fixations and explanations. The model is developed on three public datasets using a cross-dataset co-training technique.}
\label{fig:architecture}
\end{figure*}
%-------------------------------------------------------------------------
GazeXplain is built on top of a general vision-language encoder~\cite{sounak:2023:gazeformer,yinhan:2019:roberta}: Given an image (\ie, the visual stimuli) and a language instruction (\ie, the task context) as inputs, the encoder computes the image features $\Vec{V}_T\in \mathbb{R}^{d\times hw}$, the semantic embedding $\Vec{t}_I\in \mathbb{R}^{d_\text{text}}$, and the joint visual-semantic embedding $\Vec{V}_I\in \mathbb{R}^{d\times hw}$, where $h\times w$ is the size of the image feature maps, $d_\text{text}$ is the semantic embedding dimensionality, and $d$ is the joint embedding dimensionality. 

Our \textbf{attention-language decoder} employs these features in predicting explanations alongside fixations, leveraging a novel \textbf{semantic alignment} mechanism to ensure that explanations accurately reflect fixated information. GazeXplain's language input generalizes it to a wider range of eye-tracking tasks, allowing it to be trained on various eye-tracking datasets with different task designs. This \textbf{cross-dataset co-training} equips GazeXplain with a broader range of knowledge across different tasks and prevents overfitting to specific datasets, improving model robustness and generalizability.

\subsubsection{Attention-Language Decoder.} 
GazeXplain goes beyond conventional scanpath models by introducing a novel attention-language decoder to bridge the gap between visual attention patterns and natural language explanations. 

The \textbf{attention decoder} utilizes a transformer model to generate feature vectors $\{\Vec{s}_{k} | \Vec{s}_{k} \in \mathbb{R}^{d}\}_{k=1}^{K}$, indicating salient features at each temporal step, where $K$ is the maximum number of fixations. With a cross-attention mechanism, it computes the cosine similarity between $\Vec{s}_{k}$ and the joint vision-language embedding $\Vec{V}_I$ to predict the spatiotemporal distribution of fixations, denoted as $\{\Vec{m}_{k} | \Vec{m}_k\in \mathbb{R}^{h\times w}\}_{k=1}^{K}$. Additionally, it predicts parameters $\{\mu_{k}, \sigma_k^2\}_{k=1}^{K}$ characterizing the log-normal distribution of fixation durations, along with a binary indicator $\{e_k\}_{k=1}^{K}$ denoting the end of the scanpath. Following ~\cite{xianyu:2021:vqa}, we perform Monte Carlo sampling to obtain fixation positions $\{y_k\}_{k=1}^{K'}$ and durations $\{\tau_k\}_{k=1}^{K'}$, where $K'$ is the length of the sampled scanpath.

The \textbf{language decoder} in GazeXplain is a novel and distinguishing component designed to provide comprehensive semantic explanations for fixated regions, accomplished through three key steps: 
\begin{enumerate}
    \item From the visual encoder's output $\Vec{V}_T$, it extracts the local features according to the fixation position $y_k$, which results in the fixated features $\Vec{g}_k \in \mathbb{R}^{d}$ that captures the salient information within the fixated region, emphasizing localization over the entire image.
    \item To integrate visual features $\Vec{g}_k$ and semantic embedding $\Vec{t}_I$ effectively, we transform them into the same dimensionality $\Vec{g}_k^d \in \mathbb{R}^{d_\text{text}} =\Vec{W}_{d}\Vec{g}_k+\Vec{v}_I$ and $\Vec{t}_I^d\in \mathbb{R}^{d_\text{text}} =\Vec{t}_I+\Vec{v}_T$, through learnable parameters $\Vec{W}_{d}\in \mathbb{R}^{d_\text{text}\times d}$ and positional encodings $\Vec{v}_I, \Vec{v}_T\in \mathbb{R}^{d_\text{text}}$, allowing for the integration of both visual and textual information. This integration facilitates the description of local visual information in the context of task instruction.
    \item To generate the description, the features $\Vec{g}_k^d$ and $\Vec{t}_I^d$ are stacked and fed into a pre-trained language model (\eg, BLIP~\cite{junnan:2022:blip}), leveraging its contextual understanding and linguistic capabilities. This enables the generation of detailed and informative explanations $\{\Vec{w}_\ell^k\}_{\ell=1}^L$ for each fixation, where $L$ represents the length of the generated explanation. 
\end{enumerate}
By integrating visual and semantic features and incorporating language models, our language decoder enables the explanations of the scanpath predictions.

\subsubsection{Semantic Alignment.}
We propose a semantic alignment mechanism to ensure the semantic consistency between predicted fixations, explanations, and visual features. It operates by computing the cosine similarity $S_{\cos}(\cdot, \cdot)$ of different categories of features between the $i$-th and the $j$-th fixations of a scanpath:
\begin{enumerate}
    \item The \textbf{visual similarity} serves as pseudo labels for supervising the semantic alignment. It is computed as $s^r_{i,j} = S_{\cos}(\Vec{r}_i, \Vec{r}_j)$, where $\Vec{r}_i$ and $\Vec{r}_j$ represent the local image features at the fixation points, obtained from a pre-trained and frozen ResNet~\cite{kaiming:2016:resnet} model. 
    \item The \textbf{explanation similarity}, computed as $s^e_{i,j} = S_{\cos} (\Vec{e}_i^p,\Vec{e}_j^p)$, measures how closely the explanations of different fixations resemble each other semantically, where $\Vec{e}_i^p$ and $\Vec{e}_j^p$ represents the language features of the corresponding explanations, obtained from the language decoder.
    \item The \textbf{fixation similarity}, computed as $s^f_{i,j}=S_{\cos}(\Vec{g}_i,\Vec{g}_j)$, compares the fixated features acquiring global image information from the visual encoder. It measures whether the two fixations focus on similar visual information.
    \item The \textbf{multimodal similarity}, computed as $s^m_{i,j}=S_{\cos}(\Vec{e}_i^p,\Vec{g}_j)$, measures the gap between the language features $\Vec{e}_i^p$ and the visual features $\Vec{g}_j$, evaluating how well the explanations align with the visual information fixated upon.
 \end{enumerate}
Based on the similarity measures, the semantic alignment loss is denoted as
%------------------------------------------------------------------
\begin{equation}
\mathcal{L}_\text{aln} = \frac{1}{K'^2}\sum_{i=1}^{K'}\sum_{j=1}^{K'}\Big((s^e_{i,j} - s^r_{i,j})^2 + (s^f_{i,j} - s^r_{i,j})^2 + (s^m_{i,j} - s^r_{i,j})^2 \Big),
\label{equ:semantic_alignment_loss}
\end{equation}
%------------------------------------------------------------------
which compares similarities $s^e_{i,j}$, $s^f_{i,j}$, $s^m_{i,j}$ against their corresponding pseudo labels $s^r_{i,j}$. Minimizing this loss during the optimization process encourages alignment of semantic representations across fixations, explanations, and visual features, ensuring consistency in the understanding of the scanpath, fostering explanations of the visual scene throughout the scanpath. Our final training objective combines this loss with a traditional scanpath prediction loss~\cite{xianyu:2021:vqa} and a language generation loss~\cite{peter:2018:butd,oriol:2015:showtell}, jointly optimizing scanpath prediction and explanation. Please refer to the Supplementary Materials for the implementation details.

\subsubsection{Cross-Dataset Co-Training.}
Prior studies commonly focus on single dataset training~\cite{wanjie:2019:iorroi,xianyu:2021:vqa,sounak:2023:gazeformer}. For example, ChenLSTM relies on external VQA models to predict scanpaths on the AiR-D dataset~\cite{shi:2020:air}, while Gazeformer targets search-related tasks offered by COCO Search-18~\cite{zhibo:2020:cocosearch}. Such model and task dependencies limit their broader applicability. To address this, we propose cross-dataset co-training, enabling models to learn from multiple datasets simultaneously. We standardize inputs across datasets, ensuring compatibility and meaningful interaction. On the one hand, images and scanpaths are scaled to $384\times 512$ resolution. 
On the other hand, task-specific instructions are structured into a standard VQA format. For example, for free-viewing, a general question ``What do you see in the image?'' is asked, while object search instructions are converted to a question ``Is there a \texttt{[search target]} in the image?'' with a binary ``yes/no'' answer. Optionally, on datasets with behavioral responses (\eg, AiR-D, COCO-Search18), the observer's answer is also added to the instruction, which helps the model to understand inter-observer variations. 
Different from general co-training techniques relying on structured input formats~\cite{peizhao:2023:uniar}, GazeXplain's free-formed input captures rich semantics for scanpath explanation, allowing the model to understand the specific contexts and goals. In this way, models can be trained with a combination of multiple datasets, unlocking their full potential in generalization across various tasks. 

\section{Experiments}
\label{sec:experiments}

We evaluate GazeXplain through comprehensive experiments: (1) performance comparison against state-of-the-art methods, (2) ablation studies to understand component contributions, (3) evaluation of generated explanations, and (4) qualitative analysis of predicted scanpaths and explanations.
Further results, analyses, and implementation details are reported in the Supplementary Materials.

\subsection{Experimental Setup}

\textbf{Datasets.}
Our experiments utilize a combination of eye-tracking datasets. AiR-D~\cite{shi:2020:air} provides insights into human gaze behavior in VQA~\cite{yash:2017:vqa,drew:2019:gqa}, capturing gaze patterns aligned with complex visual reasoning processes. OSIE~\cite{juan:2014:osie} enriches our evaluation with eye-tracking data from free-viewing scenarios, ensuring a comprehensive assessment of our model's predictive capabilities amidst multiple salient objects. COCO-Search18~\cite{zhibo:2020:cocosearch} expands our evaluation to include both target-present and target-absent scenarios. The target-present subset focuses on gaze behavior when the search target is present, while the target-absent subset assesses our model's ability to predict gaze patterns without the target. 

\textbf{Compared Models.}
We compare GazeXplain against human ground truths and a diverse range of scanpath prediction models, including SaltiNet~\cite{marc:2017:saltinet}, PathGAN~\cite{marc:2018:pathgan}, IOR-ROI~\cite{wanjie:2019:iorroi}, ChenLSTM~\cite{xianyu:2021:vqa}, Gazeformer~\cite{sounak:2023:gazeformer}. 

\textbf{Evaluation Metrics.}
We comprehensively evaluate GazeXplain using a diverse set of metrics evaluating three aspects of models: First, with established metrics, including ScanMatch (SM)~\cite{filipe:2010:scanmatch}, MultiMatch (MM)~\cite{richard:2012:multimatch}, SED~\cite{tom:2008:sed,stephan:1997:sed,lapo:2020:wave-propagation-attention}, SS~\cite{zhibo:2020:cocosearch,zhibo:2022:targetabsent,sounak:2023:gazeformer} and SemSS~\cite{zhibo:2022:targetabsent,sounak:2023:gazeformer}, we assess scanpath models' ability to accurately predict the temporal dynamics of gaze patterns. 
In addition, we aggregate the sampled fixations into a smoothed saliency map~\cite{xiangjie:2023:scandmm}, and incorporate saliency metrics, including CC~\cite{xun:2015:salicon,ming:2015:salicon}, NSS~\cite{xun:2015:salicon,ming:2015:salicon}, AUC~\cite{zoya:2019:saliencymetric}, and sAUC~\cite{zoya:2019:saliencymetric},
to assess the spatial accuracy of the prediction.
% by aggregating the generated scanpath and post-processing the fixations to get a smoothed saliency map. 
Finally, to measure the linguistic quality of the generated textual explanations, we adopt BLEU~\cite{kishore:2002:bleu}, METEOR~\cite{satanjeev:2005:meteor}, ROUGE~\cite{chinyew:2004:rouge} and CIDEr-R~\cite{ramakrishna:2015:cider,gabriel:2021:ciderr}. This comprehensive suite of metrics allows us to assess how well the model captures the temporal, spatial, and semantic accuracies in the fixations and explanations.

\subsection{Scanpath Prediction Results}

GazeXplain demonstrates remarkable spatiotemporal accuracy in scanpath prediction, consistently surpassing state-of-the-art methods across various datasets. As shown in \cref{table:sota-scanpath-rlts}, GazeXplain's promising performance in \textbf{scanpath metrics} suggests its excellence in capturing spatial, temporal, and semantic aspects of human gaze behavior. In addition, its dominance in \textbf{saliency metrics} also indicates its ability to highlight visually important image regions.
These comprehensive results suggest that GazeXplain effectively captures the underlying patterns of human visual attention across diverse tasks and datasets, demonstrating its robustness and generalizability. The performance improvements suggest the significant role of integrated attention-language decoder, semantic alignment mechanism, and cross-dataset co-training strategy in characterizing human attention dynamics, particularly in tasks requiring semantic-level cognitive processing. %In the following, we conduct an ablation study to analyze their specific impacts on scanpath prediction and explanation.

\subsection{Ablation Study for Scanpath Prediction and Explanation}
\label{sec:ablation_study}

%-------------------------------------------------------------------------
\begin{table*}[tb]
\caption{Scanpath prediction results. The best results are highlighted
in bold.}
\label{table:sota-scanpath-rlts}
\centering
\resizebox{1\textwidth}{!}{
\begin{tabular}{l|lccccc|cccc}
 \toprule
  \multirow{2.5}{*}{Dataset}  & \multirow{2.5}{*}{Method} & \multicolumn{5}{c}{Scanpath} & \multicolumn{4}{c}{Saliency}  \\
  \cmidrule(lr){3-7} \cmidrule(lr){8-11} 
  &  & SM $\uparrow$ & MM $\uparrow$ & SED $\downarrow$ & SS $\uparrow$ & SemSS $\uparrow$ & CC $\uparrow$ & NSS $\uparrow$ & AUC $\uparrow$ & sAUC $\uparrow$  \\
\midrule
\multirow{7}{*}{AiR-D~\cite{shi:2020:air}} 
    & Human & 0.403 & 0.803 & 8.110 & 0.336 & - & 0.830 & 2.328 & 0.879 & 0.797 \\
    \cmidrule(lr){2-11}
    & SaltiNet & 0.106 & 0.750 & 10.749 & 0.117 & - & -0.014& -0.021 & 0.506  & 0.502 \\
    & PathGAN & 0.151 & 0.733 & 9.407 & 0.079 & - & 0.134 & 0.280 & 0.558 & 0.503 \\
    & IOR-ROI & 0.209 & 0.795 & 8.883 & 0.176 & - & 0.342 & 0.743 & 0.708 & 0.571 \\
    & ChenLSTM & 0.350 & 0.808 & 7.881 & 0.283 & - & 0.629 & 1.727 & 0.806 & 0.702  \\
    & Gazeformer &  0.357 & 0.811 & 7.962 & 0.287 & - & 0.550 & 1.512 & 0.760 & 0.670 \\
    & GazeXplain &  \textbf{0.386} & \textbf{0.817} & \textbf{7.489} & \textbf{0.308} & - & \textbf{0.662} & \textbf{1.851} & \textbf{0.808} & \textbf{0.719} \\
\midrule
\multirow{7}{*}{OSIE~\cite{juan:2014:osie}} 
    & Human & 0.386 & 0.808 & 7.481 & 0.332	 & - & 0.903 & 2.976 & 0.912 & 0.867 \\
    \cmidrule(lr){2-11}
    & SaltiNet & 0.149 & 0.745 & 8.768 & 0.166 & - & 0.230 & 0.556 & 0.659 & 0.596\\
    & PathGAN & 0.056 & 0.744 & 9.392 & 0.135 & - & -0.091 & -0.199 & 0.448 &  0.494\\
    & IOR-ROI & 0.290 & 0.790 & 7.826 & 0.232 & - & 0.499 & 1.426 & 0.776 & 0.673\\
    & ChenLSTM & 0.377 & 0.805 & 7.244 & 0.316 & - & 0.722 & 2.488 & 0.813 & 0.770\\
    & Gazeformer & 0.372 & 0.805 & 7.298 & 0.315 & - & 0.685 & 2.308 & 0.793 & 0.739 \\
    & GazeXplain & \textbf{0.380} & \textbf{0.806} & \textbf{7.228} & \textbf{0.317} & - & \textbf{0.748} & \textbf{2.530} &  \textbf{0.839} & \textbf{0.786} \\
\midrule
\multirow{7}{*}{\makecell{COCO-\\Search18 \\Target-\\Present~\cite{zhibo:2020:cocosearch}}}
    & Human & 0.427 & 0.810 & 1.957 & 0.510 & 0.401 & 0.861 & 3.675 & 0.944 & 0.836 \\
    \cmidrule(lr){2-11}
    & SaltiNet & 0.127 & 0.715 & 3.827 & 0.269 & 0.205 & 0.425 & 1.923 & 0.680 & 0.578 \\
    & PathGAN & 0.213 & 0.716 & 2.461 & 0.318 & 0.268 & 0.377 & 1.465 & 0.720 & 0.591 \\
    & IOR-ROI & 0.137 & 0.770 & 6.990 & 0.198 & 0.162 & 0.301 & 0.836 & 0.748 & 0.569 \\
    & ChenLSTM & 0.448 & 0.803 & \textbf{1.932} & 0.475 & 0.406 & 0.802 & 3.376  & 0.903 & 0.814\\
    & Gazeformer & 0.433 & 0.800 & 2.224 & 0.470 & 0.394 & 0.712 & 2.990 & 0.872 &  0.785 \\
    & GazeXplain & \textbf{0.480} & \textbf{0.807} & 1.981 & \textbf{0.541} & \textbf{0.443} & \textbf{0.809} & \textbf{3.529} & \textbf{0.915} & \textbf{0.836}\\
\midrule
\multirow{4}{*}{\makecell{COCO-\\Search18 \\Target-\\Absent~\cite{zhibo:2020:cocosearch}}} 
    & Human & 0.330 & 0.802 & 5.539 & 0.353 & 0.341 & 0.800 & 2.351 & 0.872 & 0.765	\\
    \cmidrule(lr){2-11}
    & ChenLSTM & 0.366 & 0.810 & 4.345 & 0.371 & 0.359 & 0.701 & 2.036 & 0.796 & 0.703 \\
    & Gazeformer & 0.354 & 0.812 & 4.492 & 0.366 & 0.353 & 0.632 & 1.837 & 0.774 & 0.681  \\
    & GazeXplain &\textbf{0.373} & \textbf{0.813} & \textbf{4.307} & \textbf{0.382} & \textbf{0.365} & \textbf{0.716} & \textbf{2.089} & \textbf{0.811} & \textbf{0.721} \\
\bottomrule
\end{tabular}
}
\end{table*}
%-------------------------------------------------------------------------

%-------------------------------------------------------------------------
\begin{table*}[!t]
\caption{Ablation study on AiR-D~\cite{shi:2020:air} for the proposed technical components: language decoder (EXP), semantic alignment (ALN), and cross-dataset co-training  (CT). The best results are highlighted in bold.}
\label{table:ablation-rlts}
\centering
\resizebox{1\textwidth}{!}{
\begin{tabular}{ccccccccccccc}
 \toprule
\multicolumn{3}{c}{Method} & \multicolumn{4}{c}{Scanpath} & \multicolumn{4}{c}{Saliency} & \multirow{2.5}{*}{CIDEr-R $\uparrow$} \\
  \cmidrule(lr){1-3} \cmidrule(lr){4-7} \cmidrule(lr){8-11} 
EXP & ALN & CT & SM $\uparrow$ & MM $\uparrow$ & SED $\downarrow$ & SS $\uparrow$ & CC $\uparrow$ & NSS $\uparrow$ & AUC $\uparrow$ & sAUC $\uparrow$  \\
\midrule
 &   &  & 0.337 & 0.805 & 8.197 & 0.274& 0.582 & 1.582 &  0.794 &  0.693 & 61.9 \\
\checkmark &  &  &  0.339 & 0.805 & 8.216 & 0.280& 0.614 & 1.674 & 0.806 & 0.706  & 91.9 \\
\checkmark &   \checkmark &  &  0.346 & 0.806 & 8.250 & 0.284 & 0.631 & 1.733 & 0.807 & 0.713  & 115.1\\
 & & \checkmark & 0.356 & 0.812 & 7.834 & 0.292 & 0.582 & 1.597 & 0.781 & 0.688  & 66.7\\
 \checkmark &  & \checkmark & 0.378 & \textbf{0.819} & 7.693 & 0.299 & 0.647 & 1.797 & 0.806 & 0.713  & 97.3 \\
 \checkmark & \checkmark & \checkmark & \textbf{0.386} & 0.817 & \textbf{7.489} & \textbf{0.308} & \textbf{0.662} & \textbf{1.851} & \textbf{0.808} & \textbf{0.719} & \textbf{123.1}\\
\bottomrule
\end{tabular}
}
\end{table*}
%-------------------------------------------------------------------------

Our GazeXplain features three key components: the language decoder for scanpath explanations (EXP), the semantic alignment mechanism (ALN), and the cross-dataset co-training (CT). The ablation study conducted on the AiR-D dataset, as shown in \cref{table:ablation-rlts}, reveals the role of each component and their joint impacts on the accuracy of scanpath prediction and explanation. To evaluate the linguistic quality of a baseline, we directly crop fixated image regions and describe them with a pre-trained BLIP captioner~\cite{junnan:2022:blip}. Please refer to the Supplementary Materials for ablation studies on the other datasets.

\textbf{Language Decoder.}
\cref{table:ablation-rlts} presents notable improvements achieved by integrating the language decoder into the model architecture. Even in the absence of semantic alignment, GazeXplain exhibits considerable improvements in scanpath prediction accuracy by explaining the scanpath. For instance, the inclusion of fixation-based explanations elevates the SM score from 0.356 to 0.378, which emphasizes the role of semantic comprehension in fostering precise and interpretable scanpath predictions. Compared to the off-the-shelf BLIP captioner used in the baseline, the CIDEr-R score is improved from 66.7 to 97.3, demonstrating the effects of our model design and training on individual datasets. These results suggest that by providing explanations for individual fixations, the model gains deeper insights into the underlying visual semantics, thereby refining its predictive capabilities.

\textbf{Semantic Alignment.}
The semantic alignment mechanism further improves the model's accuracy in identifying fixated visual semantics and generating coherent descriptions. Aligning the semantics of fixations with their corresponding explanations not only improves the precision of explanations, as observed in the improved CIDEr-R scores from 97.3 to 123.1, but also guides the model to produce more accurate fixations, reflected in the scanpath and saliency metrics (\eg, SM from 0.378 to 0.386, CC from 0.647 to 0.662). This indicates the importance of semantic coherence in guiding attention prediction models.

\textbf{Cross-Dataset Co-Training.}
Scanpath prediction research typically tackles individual tasks in isolation, each relying on its own dataset. However, our approach diverges by training a unified model across multiple datasets, harnessing shared knowledge and contemporary features to enhance performance. By leveraging diverse data sources, our model achieves notable improvements in performance across various datasets. For instance, we observe a substantial enhancement in the SM score (from 0.346 to 0.386) as well as CIDEr-R (from 115.1 to 123.1) This demonstrates the effectiveness of integrating diverse data sources for robust scanpath prediction and explanation.

\subsection{Scanpath Explanation Results}
We evaluate GazeXplain's explanatory capabilities through three main analyses: (1) assessing agreement with ground-truth annotations using language evaluation metrics, (2) analyzing the diversity and informativeness of explanations, and (3) examining its ability to accurately describe fixated objects. 

%-------------------------------------------------------------------------
\begin{figure}[tb]
  \begin{minipage}[!b]{.72\linewidth}
\captionof{table}{Explanation prediction results and diversity analysis. The best results are highlighted in bold.}
\label{table:language-rlts}
\centering
\resizebox{1\textwidth}{!}{
\begin{tabular}{llccccccc}
 \toprule
Dataset  & Method & B-4 & M  & R & C-R & Voc. & Len. & UnP\% \\
\midrule
\multirow{4}{*}{AiR-D~\cite{shi:2020:air}} 
 & w/o CT \& ALN &  27.6 & 20.5 & 50.1 & 91.9 & 557 & 100.8 & 30.92\\
 & w/o CT & 30.4 & 21.7 & 51.6 & 115.1 & 668 & 100.4 & 39.51\\
 & w/o ALN & 27.7 & 20.6 & 50.3 & 97.3 & 541 & 91.8 & 35.74\\
 & GazeXplain & \textbf{30.7} & \textbf{21.9} & \textbf{51.7} & \textbf{123.1} & 579 & 88.3 & 40.34\\
\midrule
\multirow{4}{*}{OSIE~\cite{juan:2014:osie}} 
 & w/o CT \& ALN & 12.4 & 16.5 & 40.2 & 23.6 & 633 & 103.4 & 42.08\\
 & w/o CT &  16.1 & 17.4 & 41.7 & 37.4 & 760 & 105.9 & 44.20\\
 & w/o ALN & 15.7 & 20.4 & 41.7 & 37.2 & 569 & 94.4 & 42.17 \\
 & GazeXplain & \textbf{16.7} & \textbf{21.1} & \textbf{42.0} & \textbf{48.6} & 614 & 90.9 & 44.76\\
\midrule
\multirow{4}{*}{\makecell{COCO-\\Search18 \\TP~\cite{zhibo:2020:cocosearch}}} 
 & w/o CT \& ALN & 23.3 & 15.4 & 52.4 & 111.2 & 304 & 27.3 & 64.67\\
 & w/o CT & 26.0 & 16.2 & 54.2 & 133.2 & 401 & 26.0 & 70.41\\
 & w/o ALN & 26.8 & 18.1 & 54.5 & 130.9 & 505 & 28.0 & 68.83\\
 & GazeXplain & \textbf{28.2} & \textbf{19.5} & \textbf{55.3} & \textbf{139.6} & 560 & 28.4 & 71.30\\
\midrule
\multirow{4}{*}{\makecell{COCO-\\Search18 \\TA~\cite{zhibo:2020:cocosearch}}} 
 & w/o CT \& ALN & 15.6 & 20.9 & 43.2 & 77.0 & 514 & 35.8 & 58.35\\
 & w/o CT & 17.2 & 22.5 & 43.8 & 91.9 & 583 & 35.9 & 67.03\\
 & w/o ALN & 16.3 & 26.4 &  43.2  & 92.9 & 566 & 33.3 & 66.04\\
 & GazeXplain & \textbf{18.5}  & \textbf{27.5}  & \textbf{44.5} & \textbf{106.5} & 685 & 35.5 & 71.35 \\
\bottomrule
\end{tabular}
}
  \end{minipage}\hfill
  \begin{minipage}[!b]{.24\linewidth}
    \centering
    \includegraphics[width=1\linewidth]{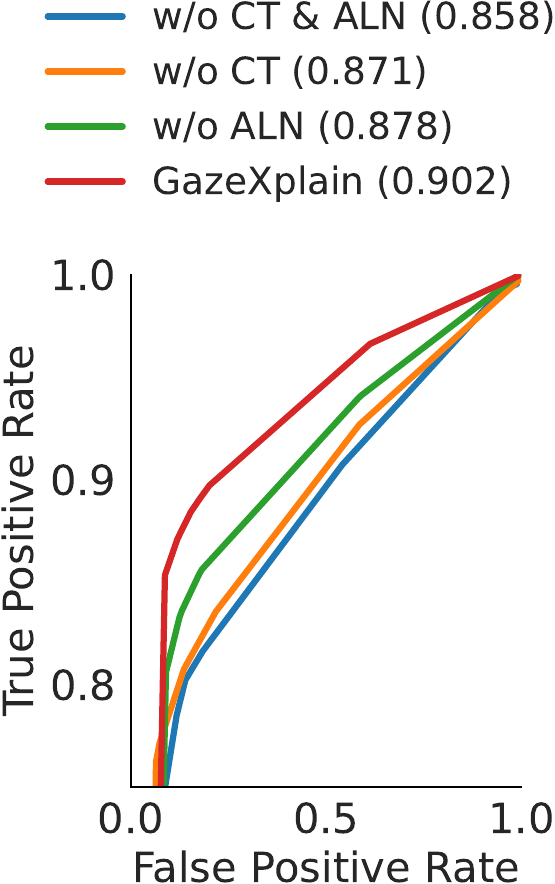}
    \caption{ROC analysis of fixations and explanations.}% \caption{Figure caption}
    \label{fig:auc-analysis}
  \end{minipage}
  
\end{figure}
%-------------------------------------------------------------------------
\textbf{Language Evaluation.}
\cref{table:language-rlts} comprehensively evaluates the agreement between generated explanations and ground-truth annotations with language metrics. GazeXplain consistently outperforms its variants (without alignment, without co-training, or both) across all datasets. The semantic alignment mechanism results in consistent performance gains across datasets (\eg, BLEU-4 from 27.7 to 30.7 and CIDEr-R from 97.3 to 123.1 on AiR-D), suggesting its significance in generating natural and fluent explanations. The co-training is more effective on OSIE (free-viewing) and COCO-Search18 (target-absent) datasets involving less structured exploration compared to the other datasets where specific objects need to be identified. It allows the model to exploit the combined information from all available data sources to learn diverse visual and linguistic relationships under these more challenging scenarios (\eg, CIDEr-R is 48.6 on OSIE, compared to the 139.6 on COCO-Search18 target-present dataset). 

\textbf{Diversity.}
To assess explanation diversity with three metrics: vocabulary size (Voc.), total explanation length per scanpath (Len.), and the percentage of unique sentences per scanpath (UnP\%). Table~\ref{table:language-rlts} reveals that incorporating semantic alignment significantly improved both vocabulary size and UnP\%. For example, on the COCO-Search18 (target-absent) dataset, vocabulary size increased from 566 to 685 words, and UnP\% increased from 66.04\% to 71.35\%. Notably, this improvement in diversity occurred while maintaining consistent explanation lengths. The COCO-Search18 dataset, known for its shorter scanpaths, naturally yielded a smaller vocabulary size, shorter explanations, and a higher percentage of unique sentences. Our co-training method, while consistently boosting UnP\%, also helped balance vocabulary sizes and explanation lengths across datasets. 
These findings highlight the importance of semantic alignment and co-training in promoting both diverse and specific explanations.

\textbf{Faithfulness.}
We evaluate the faithfulness of explanations in describing the search targets of the COCO-Search18 dataset. Specifically, we examine whether the explanation describes the search target when it is fixated on, and refrain from falsely describing it when fixations are elsewhere. To achieve this, we employ two key metrics: fixation proximity to the search target, quantified as the distance between fixations and the bounding box of the target, and semantic similarity between the generated explanation and the target, computed as the cosine distance between their embeddings using state-of-the-art techniques such as SBERT~\cite{nils:2019:sbert}. By varying spatial and semantic distance thresholds, we construct ROC curves and calculate the area under the curve (AUC) as a performance metric. Our findings, shown in \cref{fig:auc-analysis}, indicate that both semantic alignment and co-training lead to improved agreements between explanations and fixations, with AUC values increasing from 0.878 to 0.902 and 0.871 to 0.902, respectively. It suggests the significance of these techniques in aligning explanations with fixated objects. 

\subsection{Qualitative Analysis}

%-------------------------------------------------------------------------
\begin{figure}[t]
    \centering
    \includegraphics[width=1\linewidth]{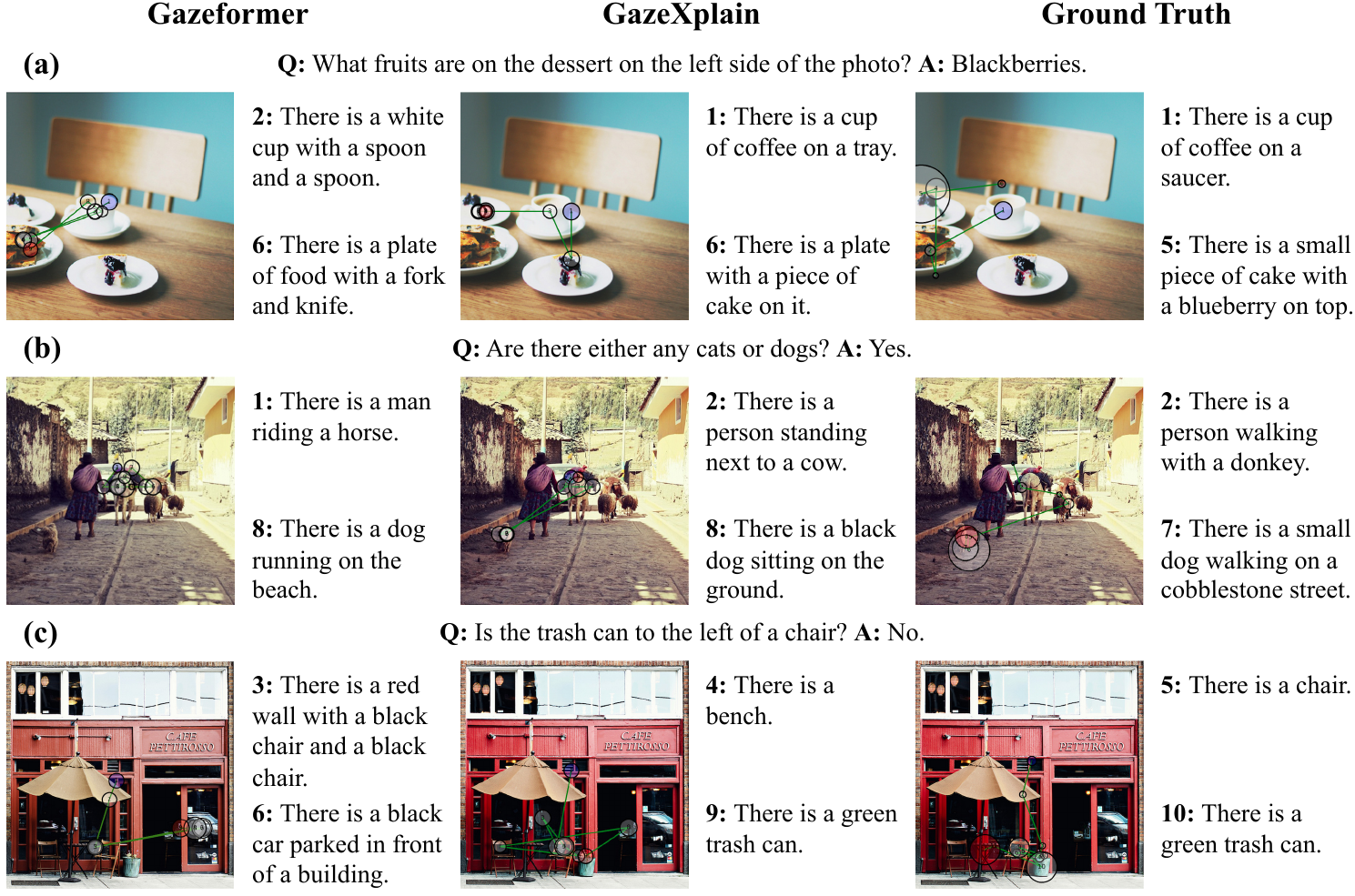}
    \caption{Quantitative examples from GazeXplain compared to Gazeformer and the ground truth. Each row shows scanpaths and  explanations of two key fixations.}
    \label{fig:qualitative_results}
\end{figure}
%-------------------------------------------------------------------------

\cref{fig:qualitative_results} presents qualitative examples of GazeXplain's scanpaths and explanations. For the Gazeformer model, we directly crop fixated image regions and describe them with a pre-trained BLIP captioner~\cite{junnan:2022:blip}. For illustration, we select two explanations describing task-relevant fixations. We observe GazeXplain's enhanced capability in predicting fixations on key objects crucial for answering questions, mirroring human gaze behavior during high-level cognitive processing. For instance, in \cref{fig:qualitative_results}a, GazeXplain accurately identifies the cake on the left side. Similarly, in \cref{fig:qualitative_results}b, the model focuses on the dog, while in \cref{fig:qualitative_results}c, it prioritizes the trash can. This alignment with human scanpaths demonstrates GazeXplain's capability of characterizing complex gaze patterns associated with cognitive tasks. Regarding explanations, while Gazeformer wrongly describes its fixations (\eg, \cref{fig:qualitative_results}a: ``a plate of food with a fork and knife'' while there is no fork or knife present, \cref{fig:qualitative_results}b: ``a man riding a horse'' while the man is walking, and \cref{fig:qualitative_results}c: ``a red wall with a black chair and a black chair'' while the chair is not black), GazeXplain provides more accurate and specific fixation descriptions. Particularly in scenes with multiple relevant objects (\eg, different types of desserts and animals in \cref{fig:qualitative_results}a-b), GazeXplain successfully distinguishes them, demonstrating robust semantic understanding. These examples illustrate GazeXplain's effectiveness in integrating visual exploration with semantic understanding, yielding more explainable and robust scanpath predictions. % The Supplementary Materials present more qualitative results.

\section{Conclusion}

We introduce GazeXplain, a novel scanpath explanation task to understand human visual attention. We provide ground-truth explanations on various eye-tracking datasets and develop a model architecture for predicting scanpaths and generating natural language explanations. The model features an attention-language decoder with a unique semantic alignment mechanism ensuring fixation-explanation consistency. Additionally, our proposed cross-dataset co-training approach enhances generalizability by leveraging diverse training datasets. Extensive experiments demonstrate GazeXplain's superior performance in both scanpath prediction and explanation, suggesting not only scanpath modeling benefits from language explanations but also GazeXplain's explanations can be integrated with other language-driven user environments. We anticipate that GazeXplain will catalyze the development of interpretable attention models, fostering advancements in human visual behavior understanding. 

\section*{Acknowledgements}
This work is supported by NSF Grant 2143197.

% \clearpage  % TODO FINAL: This \clearpage needs to be removed from both review and camera-ready versions.

\bibliographystyle{splncs04}
\bibliography{main}

\clearpage

\title{GazeXplain: Learning to Predict Natural Language Explanations of Visual Scanpaths (Supplementary Materials)} 

\titlerunning{Learning to Predict Natural Language Explanations of Visual Scanpaths}

\author{Xianyu Chen\orcidlink{0000-0002-9027-3920} \and
Ming Jiang\orcidlink{0000-0001-6439-5476} \and
Qi Zhao\orcidlink{0000-0003-3054-8934}}

\authorrunning{X.~Chen et al.}

\institute{University of Minnesota, Minneapolis MN 55455, USA\\
\email{\{chen6582,mjiang\}@umn.edu, qzhao@cs.umn.edu}}

\maketitle

\section{Introduction}
In the main paper, we have introduced GazeXplain, a novel study of visual scanpath and prediction. It involves an annotation of ground-truth explanations for diverse eye-tracking datasets related to scanpath, a general model architecture with an attention-language decoder simultaneously predicting scanpaths and the corresponding natural language explanations, a novel semantic alignment mechanism for consistent fixation-explanation alignment, and a cross-dataset co-training to generalize the scanpath prediction and explanation as well as overcome data and task-specific biases.
Our experimental results demonstrate that the proposed method achieves competitive performance and strong generalizability. The supplementary materials provide further details and additional results to support these findings:
\begin{enumerate}
   \item[1)] \cref{sec:supplementary_method} elaborates on the specific details of the proposed GazeXplain model, including the vision-language encoding module and the objective functions.
   \item[2)] \cref{sec:supplementary_implementation} presents the implementation details regarding the setting of hyperparameters and the training method of the proposed GazeXplain.
   \item[3)] \cref{sec:sup_ablation_study} presents supplementary ablation studies conducted on all three eye-tracking datasets (AiR-D~\cite{shi:2020:air}, OSIE~\cite{juan:2014:osie}, and COCO-Search18~\cite{zhibo:2020:cocosearch}). These studies evaluate the effectiveness of the three key technical components of our approach:
   \begin{itemize}
        \item Language Decoder for Scanpath Explanations (EXP)
        \item Semantic Alignment Mechanism (ALN)
        \item Cross-Dataset Co-training (CT)
   \end{itemize}
   \item[4)] \cref{sec:sup_quan} presents additional quantitative results by analyzing the generated explanations from various large vision-language models, including our GazeXplain. We provide comprehensive experiments on different prompt settings, with or without observer answers to the prompts, varied training strategies of competitors, and a more diverse range of eye-tracking datasets. These results highlight the robustness and effectiveness of our model across various scenarios.
   \item[5)] \cref{sec:sup_qual} presents additional qualitative results comparing GazeXplain's scanpaths and explanations with those generated by state-of-the-art scanpath prediction methods. These results further emphasize the superior performance of GazeXplain on the OSIE (free-viewing) and COCO-Search18 (visual search) datasets, highlighting its adaptability to various real-world visual tasks.
\end{enumerate}

\section{Supplementary Method}
\label{sec:supplementary_method}

We have introduced the novel components of our GazeXplain model architecture to address the scanpath explanation problem, including an attention-language decoder, a semantic alignment mechanism, and cross-dataset co-training. In this section, we elaborate on further details of GazeXplain's architecture, specifically focusing on the vision-language encoding process and the objective function used for training the model (as briefly mentioned in Section 3.2 of the main paper).

\subsection{Vision-Language Encoding}

GazeXplain adopts a vision encoder and a language encoder to effectively capture both the inherent visual cues within an image (bottom-up processing) and the higher-level cognitive influences stemming from the task instructions (top-down processing).

\subsubsection{Vision Encoding.}
To characterize the bottom-up stimulus-driven attention, the vision encoding involves the extraction of local image features and refining the features considering the global context:

To extract local image features, the input image is processed with a pre-trained convolutional neural network (CNN), such as the well-established ResNet-50 architecture~\cite{kaiming:2016:resnet}. The final convolutional-layer outputs of the network are extracted, denoted as $\Vec{V}_R\in \mathbb{R}^{C\times hw}$, where $C$ is the number of channels and $h$ and $w$ indicate the height and width of the feature map, respectively. The extracted features represent localized details scattered across the image, providing a foundational understanding of the visual content.

While $\Vec{V}_R$ captures localized details, it lacks a holistic understanding of the scene. To address this, GazeXplain employs a Transformer encoder~\cite{sounak:2023:gazeformer,ashish:2017:transformer,alexey:2021:vit} that excels at capturing the relationships between these local features, resulting in the refined visual features denoted as $\Vec{V}_T\in \mathbb{R}^{d\times hw}$, representing the visual content independent of the specific task at hand, where $d$ is the feature dimensionality.

\subsubsection{Language Encoding.}
Human visual attention is not solely driven by the raw visual stimuli.  GazeXplain incorporates the influence of task instructions by accepting a general task description as input. It is formatted as a question, such as ``What do you see in the image?'' or ``Is there a \texttt{[search target]} in the image?''

The task instruction is fed through a tokenizer~\cite{jonathan:1992:tokenization}, which breaks it down into a sequence of meaningful units. The tokens are then processed by a transformer-based language model, such as the powerful RoBERTa architecture\cite{yinhan:2019:roberta}. This stage generates instructional features, denoted as $\Vec{t}_I \in \mathbb{R}^{d_\text{text}}$, where $d_\text{text}$ is the feature dimensionality. Thus, the features  $\Vec{t}_I$ encapsulate the semantic meaning and intent conveyed by the task instruction.

\subsubsection{Multimodal Integration.}

Following these independent encoding stages, GazeXplain merges the bottom-up visual features $(\Vec{V}_T)$ and the top-down instructional features $(\Vec{t}_I)$ through a concatenation operation. This combined representation, denoted as $\Vec{V}_I\in \mathbb{R}^{d\times hw}$, serves as the foundation for GazeXplain's subsequent processing steps, enabling the model to leverage both visual information and task-specific guidance for accurate scanpath prediction and explanation generation.

\subsection{Objectives}

GazeXplain tackles the dual challenge of predicting scanpaths and generating explanations concurrently. To achieve this, it employs a combined loss function that guides the training process and optimizes model performance for both tasks. Given the ground-truth scanpath $\{y_k,\tau_k\}_{k=1}^{K'}$ and the language explanation $\{\Vec{w}^k\}_{k=1}^{K'}$ with the length of scanpath $K'$, where $y_k$ indicates the fixation position, $\tau_k$ indicates its duration, and $\Vec{w}^k$ is its corresponding explanation, the final training objective is a combined loss function to optimize for both scanpath prediction and explanation
%------------------------------------------------------------------
\begin{equation}
\mathcal{L} = \mathcal{L}_\text{fix} + \mathcal{L}_\text{exp} + \mathcal{L}_\text{aln},
\label{equ:total_loss}
\end{equation}
%------------------------------------------------------------------
where $\mathcal{L}_\text{fix}$ is the standard scanpath prediction loss, $\mathcal{L}_\text{exp}$ is the standard language prediction loss, and $\mathcal{L}_\text{aln}$ is the semantic alignment loss as detailed in Section 3.2 of the main paper, which encourages the model to ensure that the generated explanations exhibit a strong semantic connection with the visual features associated with each fixation. 
By carefully balancing these loss terms during training, GazeXplain not only predicts scanpaths accurately but also generates explanations that illuminate the rationale behind those fixations.

\subsubsection{Scanpath Prediction Loss.}

Given the ground truth scanpath $\{y_k,\tau_k\}_{k=1}^{K'}$, and the corresponding duration parameters $\{\mu_k,\sigma_k^2\}_{k=1}^{K'}$ of log-normal distribution from the output of GazeXplain, the scanpath prediction loss is defined as
\begin{equation}
    \mathcal{L}_\text{fix} =-\sum_{k=1}^{K'+1}\log p_k^y(y_k|y_1,\cdots,y_{k-1};\Vec{\theta}) - \sum_{k=1}^{K'}\log p_k^{\tau}(\tau_k|\mu_k,\sigma_k^2;\Vec{\theta}),
\end{equation} 
where $\Vec{\theta}$ represents the learnable parameters of GazeXplain, $\log p_k^y$ is the parametric conditioned probability of fixation position $y_k$, and $\log p_k^{\tau}$ is the parametric log-normal function~\cite{xianyu:2021:vqa}. This standard scanpath prediction loss term acts as a guiding force, encouraging the model to predict fixations that closely resemble the actual sequence of fixations observed in the ground truth data.

\subsubsection{Language Prediction Loss.}

This standard language prediction loss term ensures that the generated explanations are not only grammatically correct but also semantically consistent with the predicted scanpath and the provided task instruction.
\begin{align}
    \mathcal{L}_\text{exp} &= \frac{1}{LK'}\sum_{k=1}^{K'}\sum_{\ell=1}^{L}-\log p(\Vec{w}_{\ell}^k|\Vec{g}_k^d, \Vec{t}_I^d,\Vec{w}_{0:\ell-1}^k;\Vec{\theta}),
\end{align}
where $\Vec{\theta}$ represents the learnable parameters of GazeXplain, $\Vec{g}_k^d$ and  $\Vec{t}_I^d$ represents the encoded integration of visual and textual information mentioned in Section 3.2 of the main paper, $\Vec{w}^k$ is the ground truth language explanation of the $k$-th fixation with length $L$ and $\Vec{w}^k_\ell$ represent the $\ell$-th token of the explanation $\Vec{w}^k$. This loss term promotes the generation of explanations that accurately reflect what the model sees at each fixation point.

\section{Implementation Details}
\label{sec:supplementary_implementation}
We adhere to the original dataset splits~\cite{xianyu:2021:vqa,sounak:2023:gazeformer,zhibo:2022:targetabsent}, maintaining consistency with prior research. During training, we conduct supervised learning for 8 epochs using the Adam~\cite{diederik:2015:adam} optimizer with specific hyperparameters: a learning rate of $4\times10^{-4}$, weight decay of $5\times10^{-5}$, and batch size of 16. Subsequently, we integrate self-critical sequence training (SCST)~\cite{xianyu:2021:vqa,steven:2017:scst} for the remaining 2 epochs to enhance the model's ability to sample scanpaths and generate explanations. In SCST, the learning rate linearly decays from $10^{-5}$, with a batch size of 8, facilitating further refinement of the model's performance. The minimum and maximum lengths of the fixations for the generated scanpath are set to 1 and 16, respectively. All compared models are adapted following the same settings for fairness~\cite{xianyu:2021:vqa}.

%-------------------------------------------------------------------------
\begin{table*}[tb]
\caption{Ablation study for the proposed technical components: language decoder (EXP),  semantic alignment (ALN), and cross-dataset co-training  (CT). The best results are highlighted in bold.}
\label{table:sup-more-ablation-rlts}
\centering
\resizebox{1\textwidth}{!}{
\begin{tabular}{lccccccccccccccccc}
 \toprule
  \multirow{2.5}{*}{Dataset}  & \multicolumn{3}{c}{Modules} & \multicolumn{5}{c}{Scanpath} & \multicolumn{4}{c}{Saliency} & \multicolumn{4}{c}{Explanation $\uparrow$} \\
  \cmidrule(lr){2-4} \cmidrule(lr){5-9} \cmidrule(lr){10-13}  \cmidrule(lr){14-17} 
  & EXP & ALN & CT & SM $\uparrow$ & MM $\uparrow$ & SED $\downarrow$ & SS $\uparrow$ & SemSS $\uparrow$  & CC $\uparrow$ & NSS $\uparrow$ & AUC $\uparrow$ & sAUC $\uparrow$ & B-4 & M  & R & C-R  \\
\midrule
\multirow{6}{*}{AiR-D~\cite{shi:2020:air}} 
 &  &   &  & 0.337 & 0.805 & 8.197 & 0.274 & - & 0.582 & 1.582 & 0.794 & 0.693 & 19.5 & 18.5 & 45.0 & 61.9 \\
 & \checkmark &  &  &  0.339 & 0.805 & 8.216 & 0.280 & - & 0.614 & 1.674 & 0.806 & 0.706 & 27.6 & 20.5 & 50.1 & 91.9 \\
 & \checkmark &   \checkmark &  &  0.346 & 0.806 & 8.250 & 0.284 & - & 0.631 & 1.733 & 0.807	& 0.713 & 30.4 & 21.7 & 51.6 & 115.1\\
 &  & & \checkmark & 0.356 & 0.812 & 7.834 & 0.292 & - & 0.582 & 1.597 & 0.781 & 0.688  & 18.6 & 18.1 & 44.4 & 66.7 \\
 &  \checkmark &  & \checkmark & 0.378 & \textbf{0.819} & 7.693 & 0.299 & - & 0.647 & 1.797 & 0.806 & 0.713 & 27.7 & 20.6 & 50.3 & 97.3\\
 &  \checkmark & \checkmark & \checkmark & \textbf{0.386} & 0.817 & \textbf{7.489} & \textbf{0.308} & - & \textbf{0.662} & \textbf{1.851} & \textbf{0.808} & \textbf{0.719} & \textbf{30.7} & \textbf{21.9} & \textbf{51.7} & \textbf{123.1}\\
\midrule
\multirow{6}{*}{OSIE~\cite{juan:2014:osie}} 
 &  &   &   & 0.364 & 0.804 & 7.588 & 0.301 & - & 0.674 & 2.272 & 0.805 & 0.754 & 13.9 & 14.2 & 38.6 & 24.0 \\
 & \checkmark &  &  &  0.366 & 0.803 & 7.561 & 0.312 & - & 0.701 & 2.380 & 0.824 & 0.768 & 12.4 & 16.5 & 40.2 & 23.6 \\
 & \checkmark &   \checkmark &  &  0.369 & 0.804 & 7.633 & 0.315 & - & 0.728 & 2.414 & 0.826	 & 0.769 & 16.1 & 17.4 & 41.7 & 37.4\\
 &  & & \checkmark & 0.358 & 0.804 & 7.431 & 0.305 & - & 0.682 & 2.304 & 0.807 & 0.755 & 13.7 & 14.2 & 39.0 & 26.2 \\
 &  \checkmark &  & \checkmark & 0.372 & 0.805 & 7.392 & 0.314 & - & 0.730 & 2.471 & 0.829 & 0.776 & 15.7 & 20.4 & 41.7 & 37.2\\
 &  \checkmark & \checkmark & \checkmark & \textbf{0.380} & \textbf{0.806} & \textbf{7.228} & \textbf{0.317} & - & \textbf{0.748} & \textbf{2.530} & \textbf{0.839} & \textbf{0.786} & \textbf{16.7} & \textbf{21.1} & \textbf{42.0} & \textbf{48.6}\\
\midrule
\multirow{6}{*}{\makecell{\makecell{COCO-\\Search18 \\Target-\\Present~\cite{zhibo:2020:cocosearch}}}}
 & &  &  & 0.415 & 0.791 & 2.043 & 0.477 & 0.387 & 0.662 & 2.859 & 0.864 & 0.772 & 22.0 & 19.4 & 48.6 & 69.9 \\
 & \checkmark &  &   & 0.433 & 0.795 & 2.122 & 0.499 & 0.407 & 0.718 & 3.074 & 0.891 & 0.808 & 23.3 & 15.4 & 52.4 & 111.2\\
 &\checkmark &   \checkmark &  &  0.449 & 0.798 & 1.983 & 0.513 & 0.424 & 0.772 & 3.298 & 0.908 & 0.827 & 26.0 & 16.2 & 54.2 & 133.2\\
 &  &  & \checkmark & 0.419  & 0.800 & 2.216 & 0.487 & 0.385 & 0.675 & 2.887 & 0.874 & 0.777 & 22.4 & 19.0 & 48.1 & 67.6 \\
 & \checkmark  &  & \checkmark & 0.476 & \textbf{0.809} & \textbf{1.966} & 0.535 & 0.440 & 0.804 & 3.503 & 0.913 & 0.831 & 26.8 & 18.1 & 54.5 & 130.9\\
 &  \checkmark & \checkmark & \checkmark & \textbf{0.480} & 0.807 & 1.981 & \textbf{0.541} & \textbf{0.443} & \textbf{0.809} & \textbf{3.529} & \textbf{0.915} & \textbf{0.836} & \textbf{28.2} & \textbf{19.5} & \textbf{55.3} & \textbf{139.6}\\
\midrule
\multirow{6}{*}{\makecell{COCO-\\Search18 \\Target-\\Absent~\cite{zhibo:2020:cocosearch}}} 
 & &  & &  0.328 & 0.801 & 4.430 & 0.342 & 0.338 & 0.628 & 1.737 & 0.779 & 0.680 & 10.2 & 12.8 & 39.7 & 61.8 \\
  &\checkmark &  &  & 0.342 & 0.806 & 4.489 & 0.352 & 0.345 & 0.682 & 1.891 & 0.804 & 0.706 & 15.6 & 20.9 & 43.2 & 77.0\\
 & \checkmark &   \checkmark & &  0.349 & 0.810 & 4.409 & 0.362 & 0.354 & 0.692 & 1.948 & 0.805 & 0.711 & 17.2 & 22.5 & 43.8 & 91.9 \\
 &  & &  \checkmark &0.345 & 0.805 & 4.414 & 0.359 & 0.340 & 0.609 & 1.739  & 0.772 & 0.680 & 10.2 & 12.7 & 39.6 & 62.2 \\
 & \checkmark & & \checkmark & 0.368 & 0.811 & \textbf{4.282} & 0.378 & 0.362 & 0.704 & 2.055 & 0.802 & 0.712 & 16.3 & 26.4 & 43.2 & 92.9\\
 &  \checkmark & \checkmark & \checkmark &\textbf{0.373} & \textbf{0.813} & 4.307 & \textbf{0.382} & \textbf{0.365} & \textbf{0.716} & \textbf{2.089} & \textbf{0.811} & \textbf{0.721} & \textbf{18.5} & \textbf{27.5} & \textbf{44.5} & \textbf{106.5}\\
\bottomrule
\end{tabular}
}
\end{table*}
%-------------------------------------------------------------------------

\section{Supplementary Ablation Study}
\label{sec:sup_ablation_study}
In Tab. 3 of the main paper, we have conducted a comprehensive ablation study on the AiR-D~\cite{shi:2020:air} dataset to demonstrate the effectiveness of three key components of our proposed GazeXplain: language decoder (EXP), semantic alignment (ALN), and cross-dataset co-training (CT). In this section, to demonstrate the generalizability of our GazeXplain model and provide further insights into the contributions of these components, we conduct comprehensive ablation studies on all datasets: AiR-D\cite{shi:2020:air},  OSIE~\cite{juan:2014:osie} and COCO-Search18~\cite{zhibo:2020:cocosearch} (see \cref{table:sup-more-ablation-rlts}). Similar to the findings reported in Section 4.3 of the main paper, these results show that EXP, ALN, and CT play complementary roles in significantly enhancing overall performance on our GazeXplain:

\subsubsection{Language Decoder.}
Across all datasets, incorporating the language decoder yields significant improvements in scanpath prediction, spatial saliency, and explanation quality. This highlights the importance of explaining fixations for the model to gain a deeper understanding of the underlying visual semantics, leading to more refined predictions. In particular, when co-training is applied, there is a consistent improvement in the SM scores (0.01+ on OSIE and 0.02+ on all datasets) and CIDEr-R scores (11.0 on OSIE and 30.0+ on the other datasets). Similarly, SS, SemSS, CC, NSS and \etc scores all see a substantial increase on all the datasets, indicating that explanations enhance the model's ability to not only predict fixations accurately but also describe them in a way that is consistent with human understanding.

\subsubsection{Semantic Alignment.}
Including semantic alignment further enhances performance. We observe improvements in most metrics on all the datasets, indicating that aligning the semantics of fixations with their explanations improves both the precision of explanations and the accuracy of fixations. Across all datasets, semantic alignment yields a boost in CIDEr-R scores (about 10.0+ on all the datasets) and an improvement on almost the scanpath and saliency metric across all the datasets (0.018 increase of CC on OSIE dataset). This suggests that ensuring semantic coherence between fixations and their corresponding descriptions not only improves the quality of the explanations themselves but also guides the model to generate more accurate fixations.

\subsubsection{Cross-Dataset Co-Training.}
Co-training the model across diverse datasets consistently improves performance. This is evident from the overall increase in scores across all metrics on most datasets. Co-training allows the model to leverage complementary information from various data sources, leading to more robust scanpath prediction and explanation generation. For instance, on the COCO-Search18 Target-Present dataset, co-training results in significant improvements in both scanpath prediction (SM increases from 0.449 to 0.480) and explanation quality (CIDEr-R increases from 133.2 to 139.6). This highlights the effectiveness of co-training in enhancing the model's generalizability.

Overall, the ablation study highlights the effectiveness of each core component in GazeXplain. Language decoding empowers explanation, semantic alignment fosters coherence, and cross-dataset co-training promotes generalizability. By incorporating all three components, GazeXplain achieves superior performance in scanpath prediction, saliency prediction, and explanation generation across diverse datasets.

\section{Supplementary Quantitative Results}
\label{sec:sup_quan}

We have presented comprehensive quantitative results in the main paper, including scanpath prediction results, an ablation study of our proposed GazeXplain, and scanpath explanation results. In this section, we elaborate on further analyses and quantitative results of generated explanations from large vision-language models, explore the inclusion of observer answers during the training and inference stages, and investigate cross-dataset training strategies for competitors as well as the generalizability of GazeXplain across datasets. These analyses serve as complementary quantitative results to the main paper.

\subsubsection{Analyses on the Generated Explanations from Large Vision-Language Models.}
In the main paper, we intend to summarize the natural advantages of model-generated descriptions from large vision-language models (LVLM) over those labeled by humans, where the former is automatic, cost-effective, scalable, and possibly more consistent.To further demonstrate the quality and accuracy of the LLaVA~\cite{haotian:2023:llava} generated descriptions in the main paper, we conduct a systematic evaluation by comparing LLaVA~\cite{haotian:2023:llava} and GPT-4V~\cite{openai:2023:gpt4} descriptions of 201 red-circled COCO-Search18 objects with human annotations from Visual Genome~\cite{ranjay:2017:vg}, using CIDEr-R (C-R)~\cite{gabriel:2021:ciderr} and Sentence Similarity (SenS)~\cite{nils:2019:sbert} scores. The experimental result shows that LLaVA generates reasonably accurate descriptions (C-R=110.4, SenS=0.606), better than GPT-4V (C-R=99.1, SenS=0.592), while GazeXplain generates similarly accurate descriptions (C-R=106.3, SenS=0.590). This demonstrates that LLaVA generates more reasonable descriptions aligned with human annotations, and our GazeXplain has a similar ability to describe fixation positions by learning from the curated dataset.

This work establishes the foundation for modeling scanpath explanations by utilizing LLaVA-generated explanations. However, there are some limitations to the LLaVA-generated explanations. For example, rephrased LLaVA outputs exist due to the variability of fixations in the same region, and our manual corrections addressed outliers (less than 0.58\%).

\subsubsection{Exploration of Observer Answer.}
The AiR-D (VQA) dataset collects observers' answer during eye-tracking~\cite{shi:2020:air,xianyu:2021:vqa,xianyu:2024:individualscanpath}, which can be different from the ground-truth. This creates a new scenario for training scanpath models to be aware of task performance.
As shown in \cref{table:ablation-answer-rlts}, GazeXplain can flexibly handle different scenarios w/ or w/o observer answers: 
\begin{enumerate*}
\item When a particular observer's answer is present, it predicts the observer's scanpaths.
\item When the answer is absent, it predicts general scanpaths.
\end{enumerate*}
The main paper presents the first scenario, where SM=0.386 and NSS=1.851. Removing the answer from the test set results in a similar performance (SM=0.385, NSS=1.845). Removing the answers from both the training and test sets leads to a slight decrease (SM=0.380, NSS=1.810), but it still outperforms the compared models. This demonstrates GazeXplain's ability to capture inter-observer attention variations and provide a flexible interface for predicting either observer-specific scanpath patterns or general scanpath patterns.

%-------------------------------------------------------------------------
\begin{table*}
\caption{Ablation study on AiR-D~\cite{shi:2020:air} for the absence of observer answers in the training set and/or the test set. The best results are highlighted in bold.}
\label{table:ablation-answer-rlts}
\centering
\resizebox{1\textwidth}{!}{
\begin{tabular}{cccccccccccc}
 \toprule
\multicolumn{2}{c}{Answer Absent} & \multicolumn{4}{c}{Scanpath} & \multicolumn{4}{c}{Saliency} & \multirow{2.5}{*}{CIDEr-R $\uparrow$} \\
  \cmidrule(lr){1-2} \cmidrule(lr){3-6} \cmidrule(lr){7-10} 
Training & Test & SM $\uparrow$ & MM $\uparrow$ & SED $\downarrow$ & SS $\uparrow$ & CC $\uparrow$ & NSS $\uparrow$ & AUC $\uparrow$ & sAUC $\uparrow$  \\
\midrule
 &  & \textbf{0.386} & 0.817 & \textbf{7.489} & 0.308 & \textbf{0.662} & \textbf{1.851} & \textbf{0.808} & \textbf{0.719} & \textbf{123.1}\\
 & \checkmark &  0.385 & 0.816 & 7.539 & \textbf{0.310} & 0.659 & 1.845 & 0.805 & 0.717 & 119.6\\
 \checkmark & \checkmark & 0.380 & \textbf{0.817} & 7.684 & 0.307 & 0.653 & 1.810 & 0.801 & 0.711 & 114.4\\
\bottomrule
\end{tabular}
}
\end{table*}
%-------------------------------------------------------------------------

\subsubsection{Cross-Dataset Training for Competitors.}
To investigate whether retraining other models (ChenLSTM~\cite{xianyu:2021:vqa} and Gazeformer~\cite{sounak:2023:gazeformer}) on more datasets can improve their performance, we adjusted the settings of these models to be trained on various scanpath datasets. As shown in \cref{table:ablation-cross-dataset-rlts}, directly combining all training datasets results in lower performance compared to single-dataset training. This suggests the challenge of leveraging data from distinct tasks and settings in training. However, GazeXplain can address this challenge due to its unique model design and co-training strategy.

%-------------------------------------------------------------------------
\begin{table*}
\caption{Ablation study on the cross-dataset training strategy for all the datasets (AiR-D~\cite{shi:2020:air}, OSIE~\cite{juan:2014:osie}, and COCO-Search18~\cite{zhibo:2020:cocosearch}). The best results are highlighted in bold. $^\dagger$ indicates the model trained with the cross-dataset training strategy.}
\label{table:ablation-cross-dataset-rlts}
\centering
\resizebox{1\textwidth}{!}{
\begin{tabular}{lcccccccc}
\toprule
 \multirow{1.5}{*}{Method} & \multicolumn{4}{c}{SM $\uparrow$} & \multicolumn{4}{c}{NSS $\uparrow$}  \\
 \cmidrule(lr){2-5} \cmidrule(lr){6-9} 
($^\dagger$cross-dataset training) & AiR-D & OSIE & TP & TA & AiR-D & OSIE & TP & TA\\
\midrule
ChenLSTM $^\dagger$ & 0.325 &  0.344 & 0.358 & 0.333 & 1.790 &  2.406 & 2.694  & 1.819 \\ 
Gazeformer $^\dagger$ & 0.356 & 0.358 & 0.419 & 0.345 & 1.597 & 2.304 & 2.887 & 1.739\\ 
\midrule
ChenLSTM & 0.350 & 0.377 & 0.448 & 0.366 & 1.727 & 2.488 & 3.376 & 2.036 \\ 
Gazeformer & 0.357 & 0.372 & 0.433 & 0.354 & 1.512 & 2.308 & 2.990 & 1.837 \\ 
\midrule
GazeXplain $^\dagger$ &  \textbf{0.386} & \textbf{0.380} & \textbf{0.480} & \textbf{0.373} & \textbf{1.851} & \textbf{2.530} & \textbf{3.529} & \textbf{2.089}  \\ 
\bottomrule
\end{tabular}
}
\end{table*}
%-------------------------------------------------------------------------

\subsubsection{Generalizability across Datasets.}
To demonstrate the generalizability across different datasets, we also consider the COCO-FreeView~\cite{yupei:2022:cocofv} and WebSaliency~\cite{souradeep:2023:websaliency} datasets. COCO-FreeView~\cite{yupei:2022:cocofv} enlarges the scale of free-viewing eye fixations, offering a more appropriate testbed for free-viewing scenarios. WebSaliency~\cite{souradeep:2023:websaliency} extends the scope of natural image analysis to include webpage images and graphic designs, ensuring a thorough evaluation of our model's generalizability to non-natural images, which often contain a mix of text, images, logos, and banners. As shown in~\cref{table:sota-scanpath-extra-dataset-rlts}, GazeXplain consistently outperforms the competitors across all datasets, demonstrating promising performance in both scanpath metrics and saliency metrics.

%-------------------------------------------------------------------------
\begin{table*}[tb]
\caption{Scanpath prediction results on two additional datasets (COCO-FreeView~\cite{yupei:2022:cocofv} and WebSaliency~\cite{souradeep:2023:websaliency}). The best results are highlighted
in bold.}
\label{table:sota-scanpath-extra-dataset-rlts}
\centering
\resizebox{1\textwidth}{!}{
\begin{tabular}{l|lcccc|cccc}
 \toprule
  \multirow{2.5}{*}{Dataset}  & \multirow{2.5}{*}{Method} & \multicolumn{4}{c}{Scanpath} & \multicolumn{4}{c}{Saliency}  \\
  \cmidrule(lr){3-6} \cmidrule(lr){7-10} 
  &  & SM $\uparrow$ & MM $\uparrow$ & SED $\downarrow$ & SS $\uparrow$  & CC $\uparrow$ & NSS $\uparrow$ & AUC $\uparrow$ & sAUC $\uparrow$  \\
\midrule
\multirow{4}{*}{\makecell{COCO-\\FreeView\\\cite{yupei:2022:cocofv}}} 
    & Human & 0.340 & 0.814 & 12.782 & 0.325 & 0.830 & 1.998 & 0.869 & 0.719	\\
    \cmidrule(lr){2-10}
    & ChenLSTM& 0.360 & 0.827 & 12.243 & 0.351 & 0.790 & 1.879 & 0.820 & 0.692 \\
    & Gazeformer & 0.364 & 0.826 & 12.207 & 0.349 & 0.790 & 1.850 & 0.822 & 0.692\\
    & GazeXplain & \textbf{0.375} & \textbf{0.828} & \textbf{12.125} & \textbf{0.353} & \textbf{0.804} & \textbf{1.909} & \textbf{0.832} & \textbf{0.701}\\
\midrule
\multirow{4}{*}{\makecell{WebSaliency\\\cite{souradeep:2023:websaliency}}} 
    & Human & 0.331 & 0.838 & 18.858 & 0.213 & 0.819 & 1.720 & 0.842 & 0.768	\\
    \cmidrule(lr){2-10}
    & ChenLSTM & 0.302 & 0.819 & 16.927 & 0.199 & 0.746 & 1.348 & 0.775 & 0.679 \\
    & Gazeformer & 0.284 & \textbf{0.831} & 17.106 & \textbf{0.218} & 0.714 & 1.328 & 0.777 & 0.702\\
    & GazeXplain & \textbf{0.329} & 0.828 & \textbf{16.820} & 0.217 & \textbf{0.754} & \textbf{1.516} & \textbf{0.789} & \textbf{0.715}\\
\bottomrule
\end{tabular}
}
\end{table*}
%-------------------------------------------------------------------------

\section{Supplementary Qualitative Results}
\label{sec:sup_qual}

In addition to the qualitative examples presented in Fig. 5 of the main paper, we present more qualitative results, involving a thorough comparison of the Gazeformer model, GazeXplain, and human ground truth, covering a range of visual tasks based on the OSIE~\cite{juan:2014:osie}, COCO-Search18 Target-Present~\cite{zhibo:2020:cocosearch} and Target-Absent~\cite{zhibo:2020:cocosearch} datasets. GazeXplain consistently enhances the capability to predict fixations on key objects in these diverse tasks. These qualitative examples demonstrate the potential of our GazeXplain model as a promising and interpretable tool for unraveling the mechanisms of visual perception and attention.

%-------------------------------------------------------------------------
\begin{figure}[t]
    \centering
    \includegraphics[width=1\linewidth]{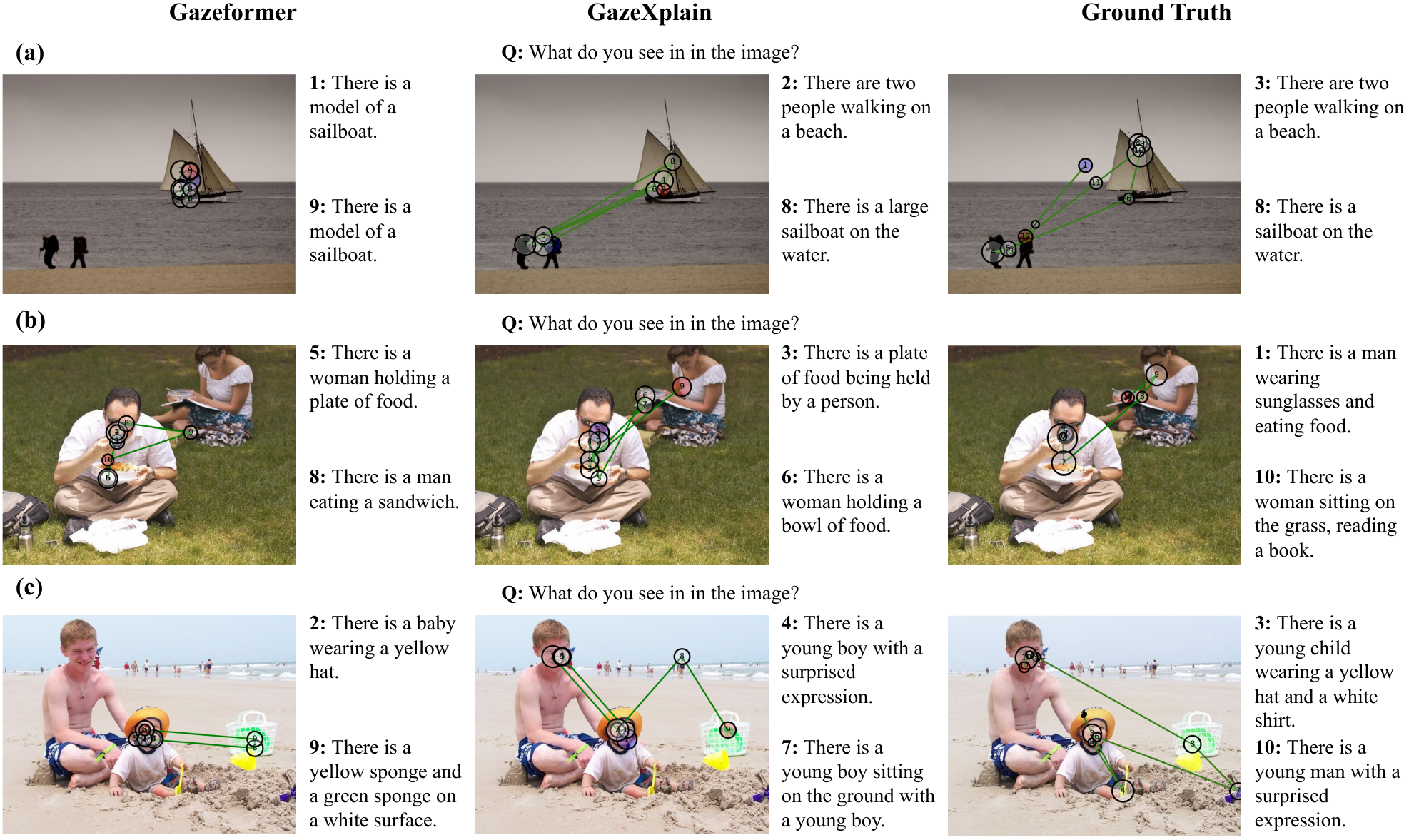}
    \caption{Quantitative examples from GazeXplain compared to Gazeformer and the ground truth on the OSIE dataset. Each row shows scanpaths and explanations of two key fixations.}
    \label{fig:sup_osie_qualitative_results}
\end{figure}
%-------------------------------------------------------------------------

\subsubsection{Results on OSIE Dataset.}

\cref{fig:sup_osie_qualitative_results} presents qualitative examples on the OSIE (free-viewing) dataset~\cite{juan:2014:osie}. Free-viewing tasks involve natural scene exploration, where observers freely gaze at a stimulus without explicit instructions. Understanding these gaze patterns is crucial for tasks like scene understanding and image retrieval. Our qualitative observations from \cref{fig:sup_osie_qualitative_results} demonstrate GazeXplain's effectiveness in free-viewing scenarios. 

We observe GazeXplain's improved ability to predict and explain fixations on salient objects. In \cref{fig:sup_osie_qualitative_results}a, GazeXplain accurately identifies the two people in the bottom-left corner, mimicking human focus on social elements within a scene. Similarly, \cref{fig:sup_osie_qualitative_results}b and \cref{fig:sup_osie_qualitative_results}c demonstrate the model's ability to detect people (a woman and a young boy) that naturally attract human attention during free-viewing. This alignment with human gaze patterns highlights GazeXplain's capability of capturing the semantic-level saliency.

Beyond fixation prediction, GazeXplain also generates accurate explanations for these fixations. Compared to Gazeformer, GazeXplain offers more precise and semantically relevant narratives. For instance, Gazeformer makes errors in all three examples: In \cref{fig:sup_osie_qualitative_results}a, it mistakenly describes a real sailboat as a ``model of a sailboat.''  Similarly, it assigns incorrect genders and objects in \cref{fig:sup_osie_qualitative_results}b and \cref{fig:sup_osie_qualitative_results}c.  In contrast, GazeXplain provides accurate descriptions, demonstrating a deeper semantic understanding of the scene.  This is particularly evident in complex scenes with multiple people (\eg,  \cref{fig:sup_osie_qualitative_results}b and \cref{fig:sup_osie_qualitative_results}c), where GazeXplain successfully distinguishes between individuals. These instances highlight GazeXplain's success in melding visual exploration with semantic insight to predict more accurate scanpaths and explanations.

%-------------------------------------------------------------------------
\begin{figure}[t]
    \centering
    \includegraphics[width=1\linewidth]{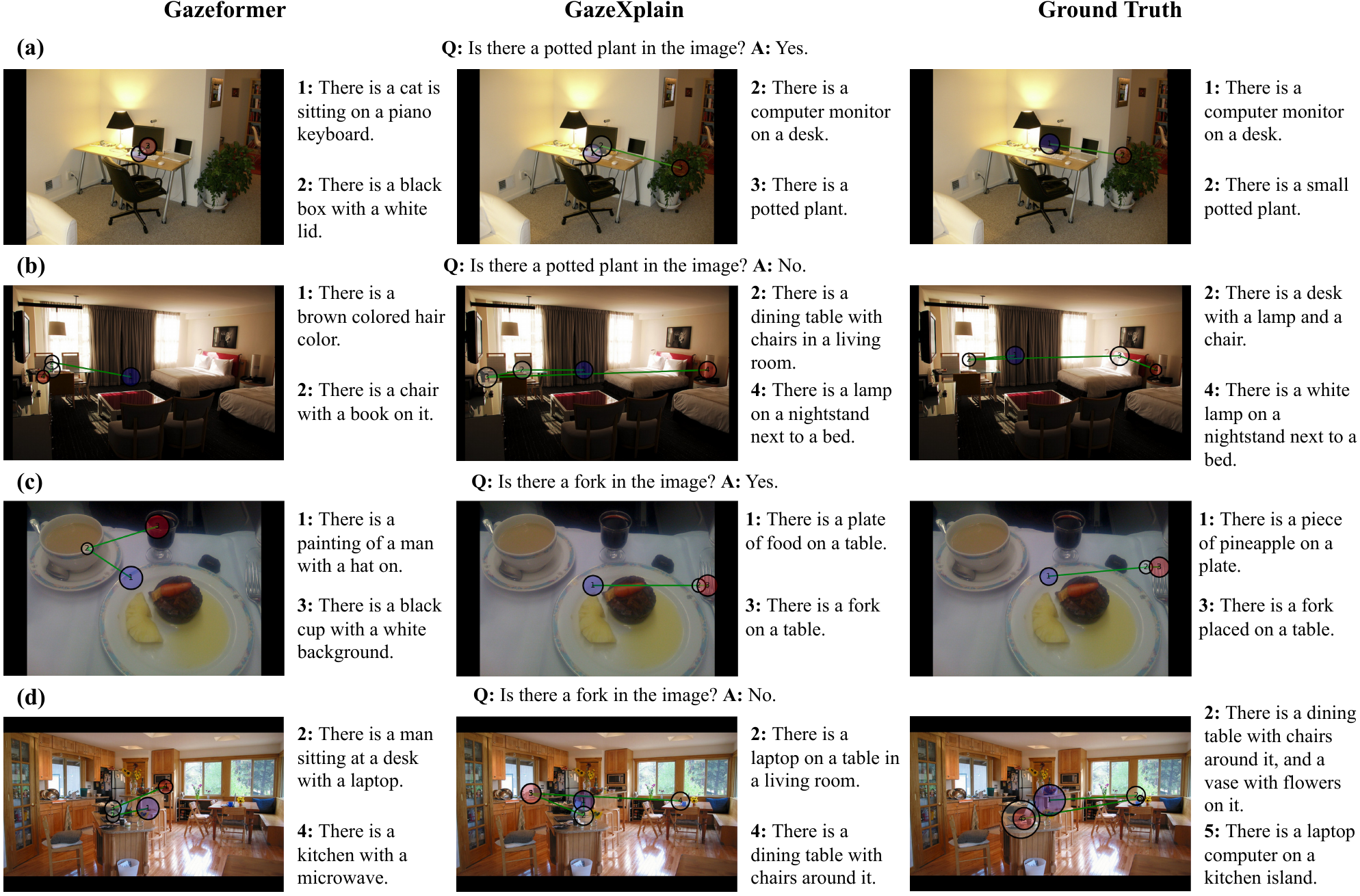}
    \caption{Quantitative examples from GazeXplain compared to Gazeformer and the ground truth on the COCO-Search18 dataset. Each row shows scanpaths and explanations of two key fixations.}
    \label{fig:sup_coco_qualitative_results}
\end{figure}
%-------------------------------------------------------------------------

\subsubsection{Results on COCO-Search18 Datasets.}

\cref{fig:sup_coco_qualitative_results} presents a qualitative comparison on the COCO-Search18~\cite{zhibo:2020:cocosearch} Target-Present and Target-Absent datasets, which feature an object search task -- finding a specific target object within an image. Our qualitative observations from \cref{fig:sup_coco_qualitative_results} demonstrate GazeXplain's effectiveness in modeling these gaze patterns.

We observe that GazeXplain accurately predicts fixations on image regions likely to contain the target object, mimicking human search strategies. For instance, when searching for a potted plant (see \cref{fig:sup_coco_qualitative_results}a and \cref{fig:sup_coco_qualitative_results}b), GazeXplain focuses on areas where a plant might typically be placed, such as the desk, floor, table, and nightstand. Similarly, in the search for a fork (see \cref{fig:sup_coco_qualitative_results}c and \cref{fig:sup_coco_qualitative_results}d), the model actively explores the table, a common location for forks.  This alignment with human search behavior highlights GazeXplain's ability to capture the cognitive process behind object search.

Beyond fixation prediction, GazeXplain's explanations are semantically aligned with the fixated objects, providing insight into the model's reasoning process.  This is in contrast to Gazeformer, which often generates inaccurate descriptions (all four examples in \cref{fig:sup_coco_qualitative_results}). For example, GazeXplain effectively explains its fixations while searching for the plant (\eg, ``desk'' in \cref{fig:sup_coco_qualitative_results}a, or ``nightstand'' in \cref{fig:sup_coco_qualitative_results}b), whereas Gazeformer makes irrelevant suggestions (\eg ``cat'' and ``piano keyboard'' in  \cref{fig:sup_coco_qualitative_results}a or ``hair'' in \cref{fig:sup_coco_qualitative_results}b). Similarly, GazeXplain offers clear explanations during the fork search (\eg, ``table'' in both \cref{fig:sup_coco_qualitative_results}c and \cref{fig:sup_coco_qualitative_results}d), while Gazeformer struggles (referring to non-existent objects, \eg, \cref{fig:sup_coco_qualitative_results}c: ``a painting of a man with a hat on'' and \cref{fig:sup_coco_qualitative_results}d: ``a man sitting at a desk with a laptop,''). These results highlight GazeXplain's capability to not only predict search fixations accurately but also to explain the rationale behind them.

\end{document}